\pdfoutput=1

\documentclass[11pt]{article}

\usepackage[final]{acl}

\usepackage{times}
\usepackage{latexsym}
\usepackage{amsmath}
\usepackage{algorithm}
\usepackage{algorithmic}
\usepackage{amsmath,amssymb,amsfonts}
\usepackage{textcomp}
\usepackage{algorithm}
\usepackage{wrapfig}
\usepackage{tabularx}
\usepackage{booktabs}
\usepackage{tcolorbox}
\usepackage{xcolor}
\usepackage{multirow}
\usepackage{makecell}
\usepackage{colortbl}
\usepackage{hyperref}
\usepackage{multicol}
\tcbuselibrary{skins}

\usepackage{microtype}

\usepackage[utf8]{inputenc}
\usepackage{inconsolata}
\usepackage{listings}
\usepackage{graphicx}
\definecolor{lightgray}{gray}{0.95}
\definecolor{deepblue}{RGB}{70,130,180}
\definecolor{deepgray}{RGB}{119,136,153}
\lstdefinestyle{prompt}{
    basicstyle=\ttfamily\fontsize{7pt}{8pt}\selectfont,
    frame=none,
    breaklines=true,
    backgroundcolor=\color{lightgray},
    breakatwhitespace=true,
    breakindent=0pt,
    escapeinside={(*@}{@*)},
    numbers=none,
    numbersep=5pt,
    xleftmargin=5pt,
    aboveskip=2pt,
    belowskip=2pt,
}
\tcbset{
  aibox/.style={
    top=10pt,
    colback=white,
    enhanced,
    center,
  }
}
\newtcolorbox{AIbox}[2][]{aibox,title={#2},#1}
\title{Can LLMs be Good Graph Judge for Knowledge Graph Construction?}

\author{Haoyu Huang$^{1}$, Chong Chen$^{2}$, Zeang Sheng$^{3}$, Yang Li$^{3}$, Wentao Zhang$^{3}$\\
  $^{1}$Hong Kong University of Science and Technology,$^{2}$Huawei Cloud BU \\
  $^{3}$Peking University \\
  \texttt{hhuangcp@connect.ust.hk,chenchong55@huawei.com} \\
  \texttt{\{shengzeang18,liyang.cs,wentao.zhang\}@pku.edu.cn}
  }

\begin{document}
\maketitle
\begin{abstract}
In real-world scenarios, most of the data obtained from the information retrieval (IR) system is unstructured. Converting natural language sentences into structured Knowledge Graphs (KGs) remains a critical challenge. We identified three limitations with respect to existing KG construction methods: (1) There could be a large amount of noise in real-world documents, which could result in extracting messy information. (2) Naive LLMs usually extract inaccurate knowledge from some domain-specific documents. (3) Hallucination phenomenon cannot be overlooked when directly using LLMs to construct KGs. In this paper, we propose \textbf{GraphJudge}, a KG construction framework to address the aforementioned challenges. In this framework, we designed an entity-centric strategy to eliminate the noise information in the documents. And we fine-tuned a LLM as a graph judge to finally enhance the quality of generated KGs. 
Experiments conducted on two general and one domain-specific text-graph pair datasets demonstrate state-of-the-art performance against various baseline methods with strong generalization abilities. Our code is available at \href{https://github.com/hhy-huang/GraphJudge}{https://github.com/hhy-huang/GraphJudge}.
\end{abstract}

\section{Introduction}
\begin{figure*}[htbp]
\centerline{\includegraphics[width=0.95\linewidth]{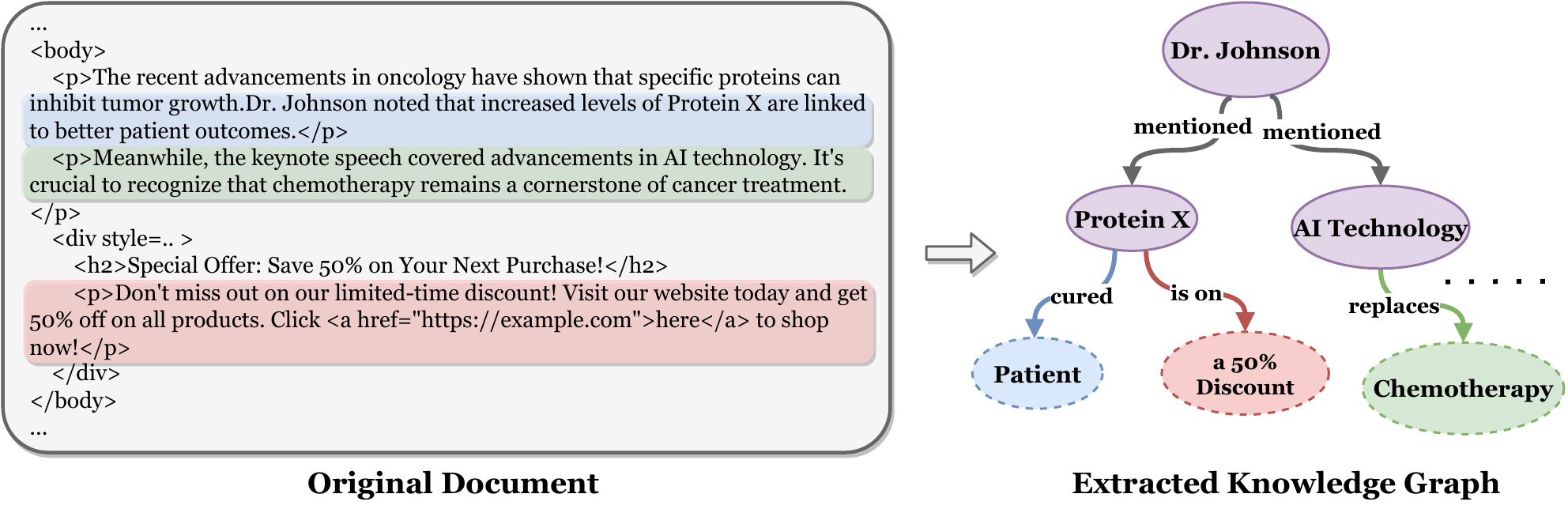}}
\caption{An demonstration of the challenges for constructing KGs with LLMs. The original document shown in the left part, while the constructed KG with some failure cases is displayed on the right side. The triple highlighted in red is wrongly formulated due to the presence of noisy information, the one in blue lacks domain knowledge, and the green-highlighted triple is a result of hallucinations by LLMs.}
\label{challenges}
\vskip -0.1in
\end{figure*}

The transition from non-structured text to structured Knowledge Graphs (KGs) is a pivotal step in the evolution of data management and information retrieval systems. The task of automatic KG construction aims to develop a structured representation of knowledge from various data sources without the need for manual intervention. KGs usually serve as the backbone of numerous data science applications, including GraphRAG systems~\cite{edge2024local,peng2024graph,huang2025retrieval} and recommendation systems~\cite{wang2019kgat,jiang2024diffkg,10904285}.
Exploring a way to construct high-quality KGs from unstructured data is crucial for different downstream applications based on KG~\cite{ge2021largeea, huang2024representation, wei2024multi, rabbani2023extraction}. 

Recently, Large Language Models (LLMs) have demonstrated significant generalization capabilities in various Natural Language Processing (NLP) tasks~\cite{pan2024unifying} and KG related tasks, such as text generation~\cite{li2024pre}, KG Completion (KGC)~\cite{yao2023exploring} and Open Information Extraction (OpenIE)~\cite{angeli-etal-2015-leveraging,dagdelen2024structured}. 
Consequently, there are many works that utilize LLMs to construct KGs from unstructured natural language documents. The incorporation of LLMs can address the issue of generalization in open-domain applications~\cite{carta2023iterative}. With its robust zero-shot generation capability, there is no need for us to gather a large volume of annotated data for tasks such as named entity recognition (NER), entity extraction, or relation extraction.


Although recent LLM-based methods~\cite{mo2025kggen,han2023pive,lairgi2024itext2kg} have gained some success in the KG construction task, we find that they may still face three challenges:

(1) \textbf{Noise Information}. Real-world documents are not only voluminous but also rife with noise, which poses a significant challenge for LLMs extracting valuable structured information. The sheer volume of data can lead to the extraction of excessive and irrelevant information, overshadowing the critical insights that LLMs are meant to uncover~\cite{liu2024much,shi2023large}. For example, as shown in Figure~\ref{challenges}, the triple \textless \textit{Protein X}, \textit{is on}, \textit{a 50\% Discount}\textgreater is incorrectly constructed due to the irrelevant advertisement with red lines in the document, which is the noise information that makes the LLM incorrectly believed that the \textit{Protein X} is on sale with discounts.

(2) \textbf{Domain-Specific Knowledge}. Naive LLMs often generate inaccurate triples with domain-specific documents, which require a deep understanding of specialized terminology and context~\cite{zhong2023comprehensive,zhu2024llms}. 
And this kind of error is hard to be observed by naive LLMs. 
For example, in Figure~\ref{challenges}, the triple \textless \textit{Protein X}, \textit{cured}, \textit{Patient}\textgreater is inaccurately extracted due to a lack of medical domain-specific knowledge. While the document marks a reference with blue lines, note that the original text only suggests a link between \textit{Protein X} and better patient results, not that it can cure patients in medical fields.

(3) \textbf{Hallucinations of LLMs}. When LLMs are directly used to build KGs, they are prone to generating false or distorted information, which is a phenomenon called hallucinations~\cite{zhang2023siren, ji2023survey}. 
This can lead to the incorporation of inaccurate or fabricated facts into the KG, undermining the reliability of the KG. For example, as shown in Figure~\ref{challenges}, the triple marked in green \textless \textit{AI Technology}, \textit{replaces}, \textit{Chemotherapy}\textgreater is incorrectly generated without any reference in the original document, even in the entity-related text highlighted with a green line.

To this end, we propose a new method called \textbf{GraphJudge}, which utilizes a fine-tuned open source LLM (e.g., LLaMA-2~\cite{touvron2023llama}) as an expert to judge the correctness of the triples generated by another closed-source LLM (e.g., GPT-4o-mini). To address the first challenge, we introduce the \textbf{Entity-Centric Text Denoising (ECTD)} module. We clean up the original documents by eliminating redundant words and irrelevant information not pertinent to the entities identified by the LLM. This module also leverages the robust zero-shot generation capabilities of LLMs to ensure the recall of a sufficient number of triple candidates~\cite{wei2023chatie, carta2023iterative}. 
To overcome the second challenge, we suggest the module of \textbf{Knowledge Aware Supervised Fine-Tuning (KASFT)}. We introduce the graph judgement task from the triple classification task. To verify the accuracy of the triples generated by the closed-source LLM, we conduct supervised fine-tuning (SFT) on an open-source LLM, which can make it achieve over 90\% accuracy on graph judgement tasks with strong generalization abilities. 
To settle the third challenge, the \textbf{Graph Judgement (GJ)} module is introduced. 
We utilize the fine-tuned open-source LLM to conduct judgement on the generated triples in the first module and filter out the wrong items to finally improve the quality of generated KGs.

In summary, the main contributions made in this work are as follows.
\begin{itemize}
    \item  Addressing challenges such as information noise, domain knowledge gaps and hallucinations in LLMs represents a critical step towards improving the quality of constructed KGs with real-world documents. To the best of our knowledge, we are the first to leverage both open- and closed-source LLMs to tackle these problems.
    \item  We propose a new framework named \textbf{GraphJudge} to leverage their capability as a graph judge and enhance the performance of LLMs in KG construction tasks. We design an entity-centric strategy to eliminate the irrelevant  and messy information in original documents. And we introduce graph judgment as the SFT task to enhance the quality of generated KGs.
    \item Experiments on two general and one domain-specific text-graph pair datasets demonstrate that GraphJudge achieves state-of-the-art performance against various baseline methods with strong generalization abilities.
\end{itemize}

\section{Related Work}
In this section, we will introduce recent LLM-based OpenIE and KG construction methods. Some work\cite{agrawal2022large, wei2023chatie} has demonstrated that LLMs have remarkable zero-shot and few-shot information extraction abilities. However, they face difficulties when it comes to more intricate tasks such as relation extraction and event extraction~\cite{carta2023iterative}. To address that, Kumar et al.~\cite{kumar2020building} propose a unified approach to construct KGs from unprocessed text. They initially fine-tuned a pre-trained language model (PLM) for NER. Subsequently, they introduced a `2-model BERT' architecture to extract relations. GPT-RE~\cite{wan2023gpt} introduces the in-context learning method and task-aware representations in demonstration retrieval and aims to enhance the connections between examples and triples. PiVe~\cite{han2023pive} designs a paradigm that fine-tuning a PLM as the verifier to predict the missing triples. With iterative verifications, the graph-based generative capability of LLMs can be improved. VicunaNER~\cite{ji2023vicunaner} utilizes the open-source LLM Vicuna to do zero-shot or few-shot NER. Similarly, it also performs recognition to identify entities that were not recognized in the previous phase. Carta et al.~\cite{carta2023iterative} develops an iterative LLM prompting-based pipeline to generate KGs without requiring predefined sets or external ontologies. iText2KG~\cite{lairgi2024itext2kg} proposes a zero-shot method to construct consistent KGs from documents with LLMs. It restructures the unprocessed documents using a preset template and identifies distinct entities and connections in a semantic manner. SAC-KG~\cite{chen2024sac} exploits LLMs as skilled automatic constructors for domain KGs and employs a naive LLM to predict the correctness of constructed triples. KGGen~\cite{mo2025kggen} clusters related entities to reduce sparsity in the KGs constructed by LLMs.
\begin{figure*}[htbp]
\centerline{\includegraphics[width=0.9\linewidth]{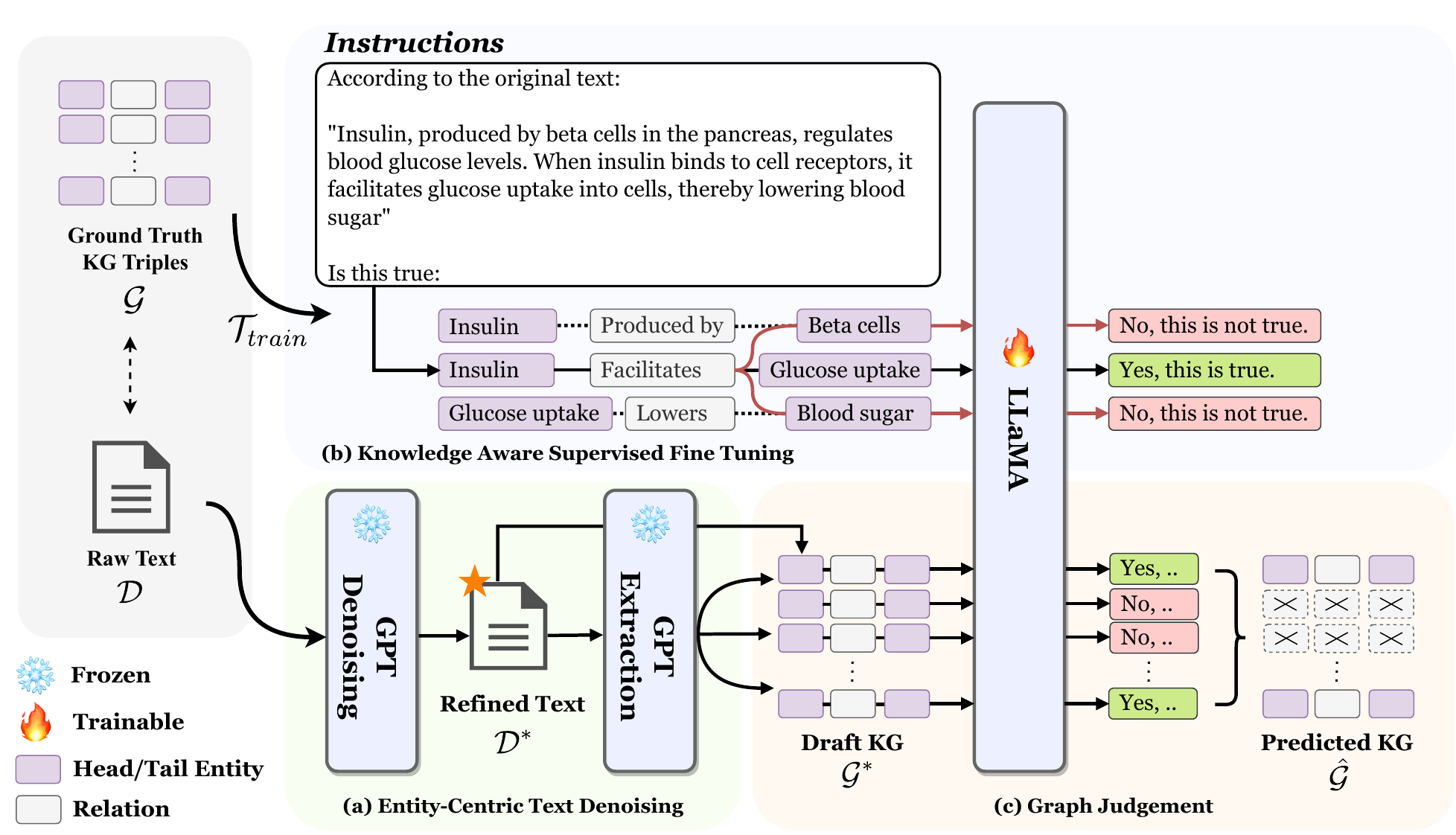}}
\caption{The overall architecture of our proposed GraphJudge framework for knowledge graph construction. It consists of three modules: (a) is the Entity-Centric Text Denoising module, (b) is the Knowledge Aware Supervised Fine Tuning module and (c) is the Graph Judgement module. The only component requiring training across the entire architecture is the open-source LLM utilized in the second module.}
\label{graphjudge}
\vspace{-0.15in}
\end{figure*}
\section{Preliminary and Definition}
In this section, we first formulate the task of KG construction and introduce the definitions we may use throughout the paper. Then we detail the definition of the graph judgement task.

\textbf{Definition 1: (Knowledge Graph Construction Task)} We define the KG construction task as a problem of how to extract entities $\mathcal{E}$ and relations $\mathcal{R}$ from a document $\mathcal{D}$, which is also called the text-to-graph generation (T2G) task. The constructed KG, is defined as $\mathcal{G} = \{(h, r, t) | h, t \in \mathcal{E}, r \in \mathcal{R}\}$, where $\mathcal{E}$ is the set of entities and $\mathcal{R}$ is the set of relations in the graph $\mathcal{G}$. In other words, each KG $\mathcal{G}$ has a corresponding original text $\mathcal{D}$. Our goal is to get a better KG $\mathcal{G}$ from a document $\mathcal{D}$.

We also define a set of KGs $\mathcal{S}_\mathcal{G} = \{\mathcal{G}_1, \mathcal{G}_2, .., \mathcal{G}_{N}\}$ and a set of documents $\mathcal{S}_\mathcal{D} = \{\mathcal{D}_1, \mathcal{D}_2, .., \mathcal{D}_{N}\}$. In our implementation, we have a set of graph-text pairs $\mathcal{S}_\mathcal{P} = \{\mathcal{P}_1, \mathcal{P}_2,.., \mathcal{P}_{N}\}$, where $\mathcal{P}_i = \{(\mathcal{G}_i, \mathcal{D}_i) | \mathcal{G}_i \in \mathcal{S}_\mathcal{G}, \mathcal{D}_i \in \mathcal{S}_\mathcal{D}\}$. And $N = |\mathcal{S}_\mathcal{P}|$ is the number of graph-text pairs.

\textbf{Definition 2: (Graph Judgement Task)} We introduce the task of graph judgement to classify each triple in generated graphs is correct or not.

Here we define the KG we constructed from a corresponding document as $\hat{\mathcal{G}}$ and $\mathcal{S}_{\hat{\mathcal{G}}}$ representing the set of graphs we constructed. And $\hat{\mathcal{T}}$ in Equation~\eqref{eq:1} represents the triples on which we need to make judgements. Our goal in the graph judgement task is to predict the label of each triple in $\hat{\mathcal{T}}$, represented as $\hat{y} \in \{0, 1\}^{|\hat{\mathcal{T}}|}$.
\begin{equation}
\hat{\mathcal{T}}=\bigcup_{\hat{\mathcal{G}} \in \mathcal{S}_{\hat{\mathcal{G}}}} \{(h,r,t) | (h,r,t) \in \hat{\mathcal{G}}\}.
    \label{eq:1}
\end{equation}
\section{Methodology}
\subsection{Overview}
As shown in Figure~\ref{graphjudge}, the proposed model \textbf{GraphJudge} consists of three modules. In the first module, which is \textbf{Entity-Centric Text Denoising}, we extract entities and relations separately following results described in \cite{Carta2023IterativeZL}. In the phrase of entity extraction, we generate entities with the denoised document. In the phrase of relation extraction, we generate relations with the entities and the denoised document as many as possible. Then, in the module of \textbf{Knowledge Aware Supervised Fine-Tuning}, we perform SFT to let the LLM become an expert in graph judgement by enhancing their abilities to check facts from documents with the triple structure and deepening their comprehension of domain-specific knowledge contained in the text-graph pairs. After that, in the final module we conduct the \textbf{Graph Judgement}. With the denoised documents as contexts, we employ the fine-tuned LLM as the graph judge to ascertain the accuracy of each triple within the graphs we generate. And then with the predicted results, we can filter out the triples that are judged as wrong. Finally, we can get high quality KGs.

\subsection{Entity-Centric Text Denoising}
\begin{figure}[htbp]
\centerline{\includegraphics[width=0.85\linewidth]{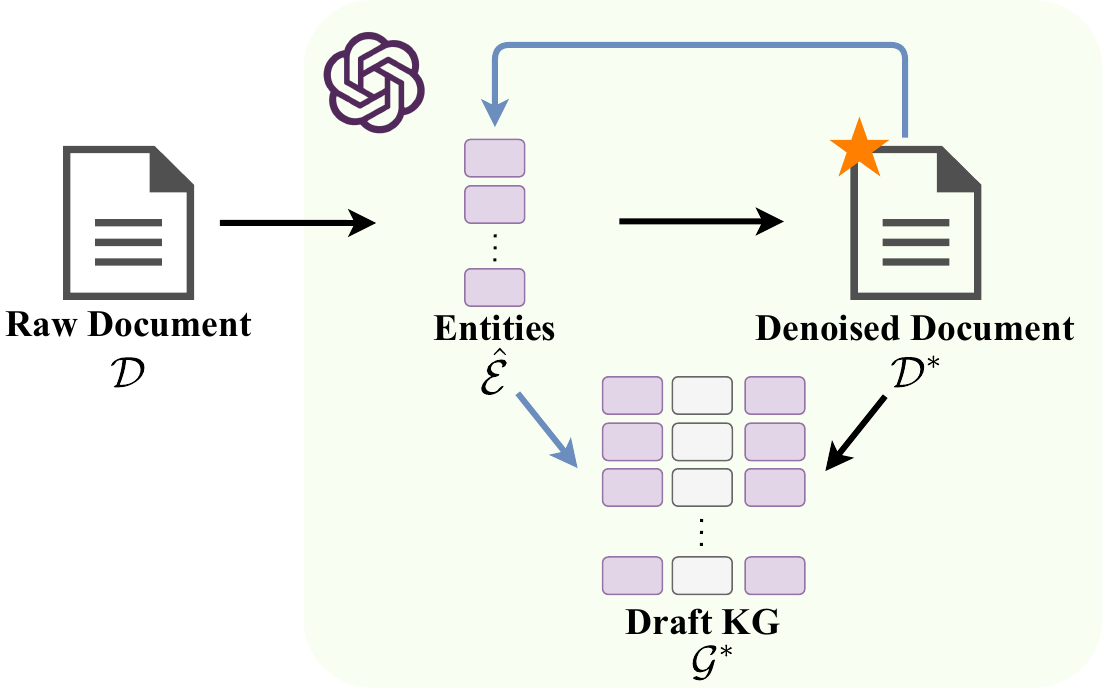}}
\caption{Illustrations of Entity-Centric Text Denoising.}
\label{fig:module1}
\vspace{-0.15in}
\end{figure}
In this module, a two-phrase extraction paradigm is designed to extract the entities and relations respectively. In phrase 1, we extract entities first and then denoise the original documents with extracted entities. In phrase 2, we conduct relation extraction and then we obtain the draft KGs. And Figure~\ref{fig:module1} is an overview of this module. In both of the two phrases we utilize a closed-source LLM to do the extraction and denoising.

\subsubsection{Text denoising and entity extraction}
In phrase 1, we consider that a substantial portion of real-world documents retrieved from information retrieval systems are consist of considerable noise information. And that may influence the quality of relations extracted by LLM~\cite{pmlr-v202-shi23a, liu2024raghelpreasoningllm}. So we design an iterative denoising method to remove messy information from the original text. 

Specifically, we extract entities from the original document using LLM. And as verified in Appendix~\ref{appendix:entity_coverage}, the entities extracted by closed-source LLMs have a high coverage rate and provide a good foundation for the following denoising, relation extraction and triple filtering processes. Subsequently, we input these entities and the original document into LLM to generate the denoised document. In this way, we can achieve two goals: (1) The noise information that is not related to the topic of the document can be removed. (2) The content of the documents can be reorganized in an entity-centric way, which is friendly to the triple extraction in the next phrase. Finally, for each raw document $\mathcal{D}$ we will get the extracted entity set $\hat{\mathcal{E}}$ and the denoised document $\mathcal{D}^{*}$. Note that important information can be well preserved in $\mathcal{D}^{*}$ as verified in Appendix~\ref{appendix:ECTD_know}.

\subsubsection{Relation extraction}
In phrase 2, we aim to extract relationships (triples) as many as feasible with the denoised document $\mathcal{D}^{*}$ and the entity set $\hat{\mathcal{E}}$ obtained in phrase 1 utilizing LLMs as shown in Equation~\eqref{eq:2}. We create numerous relationships between entities to ensure a sufficient number of suitable candidate triples for filtering with LLM judgment in the Graph Judgement module. Then we can construct a draft KG $\mathcal{G}^{*}$ for each original document $\mathcal{D}$, as illustrated in Equation~\eqref{eq:3}, where $\mathcal{R}^{*}$ is the draft relation set we generate.
\begin{equation}
\mathcal{R}^{*} = \text{LLM}(\hat{\mathcal{E}}, \mathcal{D}^{*}),
    \label{eq:2}
\end{equation}
\begin{equation}
\mathcal{G}^{*} = \{(h,r,t)|h,t \in \hat{\mathcal{E}}, r \in \mathcal{R}^{*}\}.
    \label{eq:3}
\end{equation}
\subsection{Knowledge Aware Supervised Fine-Tuning}
In this module, inspired by KG-LLaMA~\cite{yao2023exploring}, we propose the method of treating triples in the draft KG $\mathcal{G}^{*}$ as textual sequences and model graph judgement task as a sequence-to-sequence problem. We construct instruction data from the training set and fine-tune an open-source LLM to achieve the goal of both excelling at checking facts from documents with the triple structure and acknowledgment  of domain-specific knowledge. The LLM can also learn how to verify the consistency between the document and the extracted triples. 
Checking facts with the triple structure refers to the general structure of triples is often analogous to a grammatical subject, predicate, and object or a subject with a relational attribute. LLMs are anticipated to have the ability to identify their correctness from the give documents. 
Domain-specific knowledge refers to the knowledge in the documents could be a new domain~\cite{zhong2023comprehensive}, which is typically not part of pre-training data of LLMs. By employing SFT, the domain-specific knowledge from the documents can be incorporated into the LLM, thus enhances its graph judgment performance. And only if LLMs are fine-tuned as graph judges, these types of knowledge can be well learned, as justified in Figure~\ref{gj} and Appendix~\ref{appendix:KASFT}.

Before we conduct SFT on the LLM, we construct instructions for the graph judgement task with text-graph pair data. Because we need to ensure that the LLM not only excels at verifying correct triples but also skilled at telling the incorrect triples with the paired documents as contexts, we employ negative sampling to construct instruction data for training. In detail, we first sample the positive triple set $\mathcal{T}^{+}$ from the KGs of training set as described in Equation~\eqref{eq:4}, where $\mathcal{S}_{\mathcal{G}_{train}}$ is the set of all KGs in the training set.
\begin{equation}
\mathcal{T}^{+} = \bigcup_{\mathcal{G} \in \mathcal{S}_{\mathcal{G}_{train}}} \{(h, r, t^{+}) | (h,r,t^{+}) \in \mathcal{G}\}.
    \label{eq:4}
\end{equation}

Similarly, we sample negative triple set $\mathcal{T}^{-}$ from the KGs in training set as described in Equation~\eqref{eq:5}, where $\mathcal{E}$ represents the entity set of the graph $\mathcal{G}$. $(h, r, t^{-})$ is a negative triple of the graph $\mathcal{G}$, where $t^{-}$ is a negative entity. We replace the positive tail entity $t^{+}$ in each positive triple with a randomly selected negative tail entity $t^{-}$. Note that if the selected negative entity is the same as or similar to the original one, we will skip that because they may not construct a triple reflecting a false fact.
\begin{equation}
\begin{aligned}
\mathcal{T}^{-} &= \bigcup_{\mathcal{G} \in \mathcal{S}_{\mathcal{G}_{train}}} \{(h, r, t^{-}) | \\ &\quad (h,r,t^{+}) \in \mathcal{G}, t^{-} \in \mathcal{E} \setminus \{t^{+}\} \}.
\label{eq:5}
\end{aligned}
\end{equation}

Then we merge the positive triple set $\mathcal{T}^{+}$ and negative triple set $\mathcal{T}^{-}$ constructed from KGs $\mathcal{S}_{\mathcal{G}_{train}}$. Then we can obtain all the triples $\mathcal{T}_{train}$ we need to construct instructions.
\begin{equation}
\mathcal{T}_{train} = \mathcal{T}^{+} \cup \mathcal{T}^{-}.
    \label{eq:6}
\end{equation}

Furthermore, we transfer the triples in $\mathcal{T}_{train}$ to natural language sentences to construct the instruction data with paired documents $\mathcal{D}$ as contexts following the prompt templates shown in Appendix~\ref{appendix:prompt_template}. The triple sentences either represent a real fact or a fake fact. Then let the LLM make judgements with these instructions. Mathematically, with tokenized sentences $\mathcal{X}_{\mathcal{T}_{train}}$ transferred from triples $\mathcal{T}_{train}$ and paired documents $\mathcal{D}$, and tokenized instruction $\mathcal{X}_\mathcal{I}$, for a sequence of length $L$, we compute the probability of generating the target output $\mathcal{X}_\mathcal{O}$ as follows: 
\begin{equation}
p(\mathcal{X}_\mathcal{O} | \mathcal{X}_t, \mathcal{X}_\mathcal{I}) =  \prod_{i=1}^{L} p_{\theta}(x_{i} | \mathcal{X}_t, \mathcal{X}_{\mathcal{I},<i}, \mathcal{X}_{\mathcal{O},<i}),
    \label{eq:7}
\end{equation}
where $\mathcal{X}_{t} \in \mathcal{X}_{\mathcal{T}_{train}}$. And $\theta$ are the learnable parameters within the open-source LLM to be fine-tuned.

\subsection{Graph Judgement}
The KGs created in the first module are preliminary and that is also why we call that draft KGs. In this module, we will judge the factual correctness of the triples in these draft KGs using our fine-tuned LLM in the second module and filter out the incorrect triples.

In detail, we let LLM do the graph judgement task on the draft KGs $\mathcal{G}^{*}$. Here we define draft KG set as $\mathcal{S}_{\mathcal{G}^{*}}$, and the triples in all draft KGs can be symbolized as
\begin{equation}
\mathcal{T}^{*}=\bigcup_{\mathcal{G}^{*} \in \mathcal{S}_{\mathcal{G}^{*}}} \{(h,r,t) | (h,r,t) \in \mathcal{G}^{*}\}.
    \label{eq:8}
\end{equation}

Then, the LLM needs to assess the correctness of each triple in $\mathcal{T}^{*}$ by considering whether it aligns with the knowledge in paired documents and avoids conflicting with both domain-specific knowledge as it learned. We obtain the predictions of all the triples in $\mathcal{T}^{*}$ with the learned parameters $\theta$ as shown in Equation~\eqref{eq:9}. And $\text{Pred}(\cdot)$ is a function that transforms the outputs of LLM into the binary results $\hat{y} \in \{0, 1\}^{|\mathcal{T}^{*}|}$. Based on the judgments made by LLM, we filter the triples $\mathcal{T}^{*}$ in draft KGs to obtain high-quality triples $\hat{\mathcal{T}}$ as described in Equation~\eqref{eq:10}, which form the final KGs we seek. $\hat{y}_{(h,r,t)}$ is the predicted result of a triple $(h,r,t)$.
\begin{equation}
\hat{y} = \text{Pred}(p_{\theta}(\mathcal{X}_{\mathcal{T}^{*}})),
    \label{eq:9}
\end{equation}
\begin{equation}
\hat{\mathcal{T}} = \{(h,r,t) \in \mathcal{T}^{*} | \hat{y}_{(h,r,t)} = 1\}.
    \label{eq:10}
\end{equation}

Similarly, the refined relation set $\hat{\mathcal{R}} = \{r | (h,r,t) \in \mathcal{G}^{*}, \hat{y}_{(h,r,t)} = 1\}$ can also be obtained. Lastly, for each draft KG $\mathcal{G}^{*} \in \mathcal{S}_{\mathcal{G}^{*}}$ we can get the refined KG $\hat{\mathcal{G}}$ that we desire as shown in Equation~\eqref{eq:12}. The implementation details of the graph judgment procedure are demonstrated in Appendix~\ref{sec:graphjudgement_algorithm}.
\begin{equation}
\hat{\mathcal{G}} = \{(h,r,t) | h,t \in \hat{\mathcal{E}}, r \in \hat{\mathcal{R}}, (
h,r,t) \in \hat{\mathcal{T}}\}.
    \label{eq:12}
\end{equation}

\section{Experiments}

In this section, we will conduct experiments to address the following key research questions: \textbf{RQ1: }How well does GraphJudge perform on both general knowledge data and domain-specific knowledge data? \textbf{RQ2:} How do the different key components in our proposed method GraphJudge contribute to its overall performance? \textbf{RQ3:} How about the generalization capability of GraphJudge when applied across different datasets?

\subsection{Experimental Settings}

\subsubsection{Dataset}
In our study, we conduct experiments on two general datasets (\textbf{REBEL-Sub}~\cite{huguet-cabot-navigli-2021-rebel-relation} and \textbf{GenWiki}~\cite{jin-etal-2020-genwiki}) and two domain-specific datasets (\textbf{SCIERC}~\cite{luan2018multitask} and the Windows-centric subset of \textbf{Re-DocRED}~\cite{tan2022revisiting} used by \citet{sun2025lkd}) with golden ground truth KGs. We demonstrate the detailed information and statistics of each dataset in Appendix~\ref{sec:appendix_dataset}. For each dataset we randomly select a sample of 2000 data points from the training data for validation purposes during the fine-tuning of the LLM.

\subsubsection{Baselines}
In our performance comparison, we consider six baselines for comprehensive evaluation: \textbf{GPT-4o-mini}: We conduct experiments on GPT-4o with one-shot learning method. The instructions we have developed are identical to those outlined in our method. \textbf{GPT-4o}~\cite{hurst2024gpt}: The same settings as GPT-4o-mini. \textbf{RAKG}~\cite{zhang2025rakg}, \textbf{iText2KG}~\cite{lairgi2024itext2kg}, and \textbf{KGGen}~\cite{mo2025kggen}: We follow the default settings of them and their official implementations with GPT-4o-mini as the LLM. \textbf{PiVe}~\cite{han2023pive}: We follow the default parameter settings of PiVe. We use the largest verifier module in PiVe, Flan-T5-XXL~\cite{chung2024scaling}. We employ the LoRA adapter checkpoint\footnote{https://huggingface.co/Jiuzhouh/flan-t5-xxl-lora-verifier}, which has been well trained. And the LLM we use in this model is GPT-4o-mini. We implement an iterative prompting approach with three rounds, which represents the optimal number of iteration rounds as outlined in their study.

\subsubsection{Implementation Details}
\begin{table*}[ht]
\centering
\renewcommand{\arraystretch}{1.2}
\scalebox{0.6}{
\begin{tabular}{c|l|cc>{\columncolor[gray]{0.95}}c|cc>{\columncolor[gray]{0.95}}c|cc>{\columncolor[gray]{0.95}}c}
\toprule
                          \textbf{Dataset}&\textbf{Method}& \textbf{G-BS-Acc↑}& \textbf{G-BS-Recall↑}& \textbf{G-BS-F1↑}& \textbf{G-BL-Acc↑}& \textbf{G-BL-Recall↑}& \textbf{G-BL-F1↑}&\textbf{G-RO-Acc↑}& \textbf{G-RO-Recall↑}& \textbf{G-RO-F1↑}\\ \midrule
\multirow{7}{*}{REBEL-Sub} 
                          & \textit{GPT-4o-mini}& 0.3571& 0.9024& 0.4289& 0.2343& 0.6687& 0.3018& 0.2095& 0.6266& 0.2779\\
                          & \textit{GPT-4o}& 0.3131& 0.9432& 0.4163& 0.2345& 0.7284& \underline{0.3158}& 0.2201& 0.6851& \underline{0.2966}\\
                          & \textit{RAKG}& \cellcolor{red!10}{0.1196} & 0.9571 & 0.2127 & \cellcolor{red!10}{0.1078} & 0.8625 & 0.1917 & \cellcolor{red!10}{0.1012} & 0.8095 & 0.1799\\
                          & \textit{iText2KG}& 0.3847 & 0.9342 & 0.4937 & 0.2704 & 0.6579 & 0.3504 & 0.2180 & 0.5475 & 0.2864\\
                          & \textit{PiVe}& 0.3082& 0.9378& 0.4090& 0.2217& 0.7089& 0.3010& 0.2068& 0.6693&0.2823\\
                          & \textit{KGGen}& 0.4190& \cellcolor{red!10}{0.8937}& \underline{0.4995}& 0.2587& \cellcolor{red!10}{0.5794}& 0.3146& 0.2233& \cellcolor{red!10}{0.5037}& 0.2719\\
                          & \textit{GraphJudge}& 0.4868& 0.9144& \textbf{0.5796}& 0.3391& 0.6490& \textbf{0.4057}& 0.3032& 0.5878& \textbf{0.3571}\\
\midrule
\multirow{7}{*}{GenWiki} 
                        & \textit{GPT-4o-mini}& 0.7825& 0.9334& 0.8368& 0.6136& 0.7353& 0.6577& 0.5451& 0.6568& 0.5857\\
                        & \textit{GPT-4o}& 0.7871& 0.9393& \underline{0.8428}& 0.6318& 0.7561& \underline{0.6774}& 0.5614& 0.6742& \underline{0.6028}\\
                        & \textit{RAKG}& \cellcolor{red!10}{0.4695} & 0.9521 & 0.6058 & \cellcolor{red!10}{0.3804} & 0.7991 & 0.5035 & \cellcolor{red!10}{0.3397} & 0.7196 & 0.4507\\
                        & \textit{iText2KG}& 0.8984 & \cellcolor{red!10}{0.7611} & 0.7986 & 0.7193 & 0.6007 & 0.6310 & 0.6042 & \cellcolor{red!10}{0.5227} & 0.5432 \\
                        & \textit{PiVe}& 0.7463& 0.9485& 0.8230& 0.5884& 0.7516& 0.6503& 0.5251& 0.6746& 0.5817\\
                        & \textit{KGGen}& 0.8578& 0.8230 & 0.8169& 0.5799& \cellcolor{red!10}{0.4700}& 0.5542& 0.4845& 0.5598& 0.4641\\
                        & \textit{GraphJudge}& 0.7936& 0.9375& \textbf{0.8457}& 0.6407& 0.7591& \textbf{0.6836}& 0.5714& 0.6796& \textbf{0.6106}\\
\midrule
\multirow{7}{*}{SCIERC} 
                        & \textit{GPT-4o-mini}& 0.5974& 0.9183& 0.6882& 0.4368& 0.6725& \underline{0.5040}& 0.3876& 0.6065& \underline{0.4490}\\
                        & \textit{GPT-4o}& 0.6272& 0.9079& \underline{0.7035}& 0.4469& 0.6530& 0.5032& 0.3914& 0.5807& 0.4425\\
                        & \textit{RAKG}& \cellcolor{red!10}{0.2137} & 0.9528 & 0.3474 & \cellcolor{red!10}{0.1647} & 0.7334 & 0.2678 & \cellcolor{red!10}{0.1556} & 0.6864 & 0.2524 \\
                        & \textit{iText2KG}& 0.8100 & 0.6674 & 0.6724 & 0.5747 & 0.4732 & 0.4772 & 0.4836 & \cellcolor{red!10}{0.3968} & 0.3999 \\
                        & \textit{PiVe}& 0.5738& 0.9225& 0.6725& 0.4192& 0.6757& 0.4924& 0.3719 & 0.6092& 0.4385\\
                        & \textit{KGGen}& 0.8394& \cellcolor{red!10}{0.6500}& 0.6635& 0.6045& \cellcolor{red!10}{0.4594}& 0.4725& 0.5426& 0.4100& 0.4211\\
                        & \textit{GraphJudge}& 0.6847& 0.8775& \textbf{0.7283}& 0.4898& 0.6273& \textbf{0.5216}& 0.4321& 0.5591& \textbf{0.4603}\\
\midrule
\multirow{7}{*}{Re-DocRED} 
                        & \textit{GPT-4o-mini}& 0.8036 & 0.7460 & 0.6807 & 0.4840 & 0.4659 & \underline{0.4254} & 0.3337 & 0.3338 & \underline{0.2938} \\
                        & \textit{GPT-4o}& 0.7508 & 0.7588 & \underline{0.6864} & 0.4438 & 0.4682 & 0.4066 & 0.3076 & 0.3383 & 0.2833 \\
                        & \textit{RAKG}& \cellcolor{red!10}{0.3121} & 0.9243 & 0.4422 & \cellcolor{red!10}{0.2144} & 0.6377 & 0.3044 & \cellcolor{red!10}{0.1613} & 0.4922 & 0.2306\\
                        & \textit{iText2KG}& 0.7957 & 0.5519 & 0.5502 & 0.4930 & 0.3741 & 0.3522 & 0.3704 & 0.2958 & 0.2678\\
                        & \textit{PiVe}& 0.7963& 0.6769 & 0.6289 & 0.4545 & 0.4193 & 0.3729 & 0.2848 & 0.2955 & 0.2506\\
                        & \textit{KGGen}& 0.8422 & \cellcolor{red!10}{0.3914} & 0.4372 & 0.5284 & \cellcolor{red!10}{0.2594} & 0.2768 & 0.3662 & \cellcolor{red!10}{0.1937} & 0.1959\\
                        & \textit{GraphJudge}& 0.7801 & 0.7579 & \textbf{0.7051} & 0.4776 & 0.4830 & \textbf{0.4322} & 0.3350 & 0.3532 & \textbf{0.3048} \\
\bottomrule
\end{tabular}
}
\caption{Comparisons of GraphJudge with six baseline methods across four datasets. The cells marked with \colorbox{red!10}{red color} hold the worst performance in each column of Acc and Recall. The \textbf{best} and \underline{second-best} results are also highlighted in each column of F1 scores.}
\label{tab:main_results}
\vskip -0.1in
\end{table*}


\textbf{Large Language Model}: The LLMs we employed in this research are various in different modules. In the ECTD module, we utilize the closed-source LLM GPT-4o-mini to denoise the original documents and extract triples from documents. In the KASFT module, an open-source LLM LLaMA-2-7B~\cite{touvron2023llama} is used as our base model.

\textbf{Supervised Fine-Tuning}: We employ LLaMA-2-7B as the base model to carry out SFT with LoRA~\cite{hu2021lora}. The instructions are constructed with the documents, query sentences, and the triple sentences. We perform SFT on autoregression generation tasks, which is a common approach to fine-tune LLMs~\cite{black2022gpt}. The expected responses (labels) are either `Yes, that is true.' or `No, that is not true.'. Training settings are illustrated in Appendix~\ref{sec:appendix_settings}. The training was done using a single L20 GPU with 48GB of RAM.

\subsubsection{Evaluation Metrics}
We acknowledge that conventional evaluation techniques are rule-based. They assess the resemblance between predictions and ground-truth KGs through strict string matching, potentially overlooking semantic similarities. Therefore, to better evaluate the quality of the produced KGs against the ground-truth KGs, similar to PiVe~\cite{han2023pive}, we utilize one semantic level and two soft string matching evaluation metrics to calculate the \textbf{Accuracy}, \textbf{Recall}, and \textbf{F1} scores: \textbf{G-BERTScore (G-BS)}, \textbf{G-BLEU (G-BL)} and \textbf{G-ROUGE (G-RO)}. We elaborate on them in Appendix~\ref{sec:appendix_metrics}.

\subsection{Overall Performance Comparison (RQ1)}
We demonstrate the evaluation results of our method GraphJudge with GPT-4o-mini and other baseline methods across three datasets in Table~\ref{tab:main_results}. We have the following insights:

\textbf{GraphJudge's superior performance.} GraphJudge outperforms other baselines in most of the cases. The superiority of GraphJudge’s F1 scores (marked with gray color) demonstrates that, while maintaining a reasonable level of recall for triples, it also achieves improvement in accuracy. For example, as the results marked with red color show, although RAKG and PiVe exhibit stronger recall ability, they overlook triple accuracy. KGGen excels in accuracy but fails at recall. In contrast, GraphJudge leverages the ECTD module based on a closed-source LLM to ensure recall ability, while the KASFT and GJ modules with a fine-tuned open-source LLM guarantee accuracy, enabling its F1 score to surpass those of other baseline models. We can also observe that GraphJudge excels not only with domain-specific documents, but also demonstrates superior performance with general documents. 

\textbf{GraphJudge is cost-effective.} Remarkably, GraphJudge achieves state-of-the-art performance by fine-tuning only a 7B LLM, which is significantly more efficient and cost-effective compared to the 70B LLM employed in PiVe. In addition, GraphJudge can even outperform GPT-4o with GPT-4o-mini, which is a small model with lower token cost. However, other baseline methods fail to achieve that.

\subsection{Module Ablation Study (RQ2)}\label{sec:ablation}
\begin{table}[ht]
\centering
\renewcommand{\arraystretch}{1.2}
\scalebox{0.7}{\begin{tabular}{c|l|ccc}
\toprule
                         \textbf{Dataset} & \textbf{Method} & \textbf{G-BS-F1↑}& \textbf{G-BL-F1↑}& \textbf{G-RO-F1↑}\\ \midrule
\multirow{4}{*}{REBEL-Sub}& \textit{GraphJudge}& \textbf{0.5796}& \textbf{0.4057}& \textbf{0.3571}\\
                          & \textit{w/o ECTD}& 0.4548& 0.3343&0.3094\\
                          & \textit{w/o GJ}& 0.4203& 0.3052& 0.2820\\
                          & \textit{w/o KASFT}& 0.4506& 0.3219& 0.2935\\
                          \midrule
\multirow{4}{*}{SCIERC}& \textit{GraphJudge}& \textbf{0.7283}& \textbf{0.5216}& \textbf{0.4603}\\
                          & \textit{w/o ECTD}& 0.6818& 0.5029& 0.4509\\
 & \textit{w/o GJ}& 0.7172& 0.5146& 0.4552\\
                          & \textit{w/o KASFT}& 0.6700& 0.4644& 0.4084\\ \bottomrule
\end{tabular}}
\caption{The results of ablation study on REBEL-Sub dataset and SCIERC dataset.}
\label{tab:ablation}
\vskip -0.1in
\end{table}

We perform an ablation study to explore the specific impacts of various modules within GraphJudge, and the results are reported in Table~\ref{tab:ablation}. The insights are outlined below:
\begin{figure}[htbp]
\centerline{\includegraphics[width=0.95\linewidth]{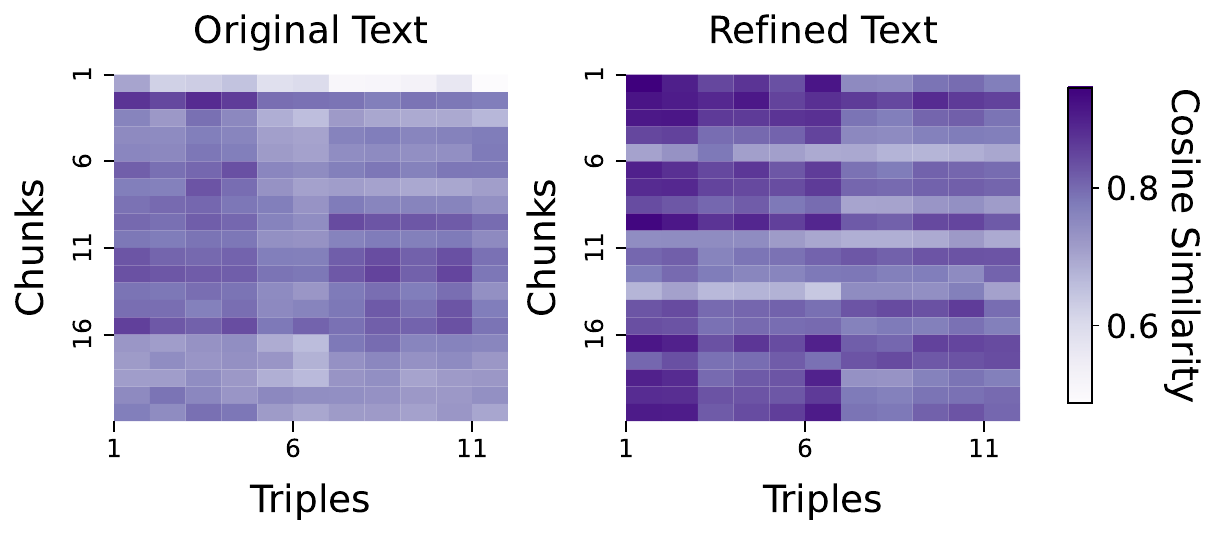}}
\caption{(a) The left map is the semantic similarity between the original document and paired KG triples. (b) The right map is the semantic similarity between the denoised document and paired KG triples.}
\label{relavance}
\vspace{-0.15in}
\end{figure}

\textbf{Effect of Entity-Centric Text Denoising}. We investigate the benefit of introducing entity-centric denoising paradigm using the variant `w/o ECTD', where we do not conduct document denoising and directly extract entities and relations from original documents. The results show that our full model performs significantly better than this ablated version. It suggests that ECTD module can avoid LLMs extract wrong structured information from irrelevance or not well-formatted corpus. 

Furthermore, to showcase the noise reduction capability of ECTD, we visualize the semantic correlation of the triples in a known KG with the denoised and original document, respectively. As shown in Figure~\ref{relavance}, deeper color in the heat maps suggests a stronger relevance. The refined document exhibits greater relevance to the triples, demonstrating the effectiveness of ECTD. Implementation details are described in Appendix~\ref{appendix:ECTD}.

\begin{figure}[htbp]
\centerline{\includegraphics[width=0.95\linewidth]{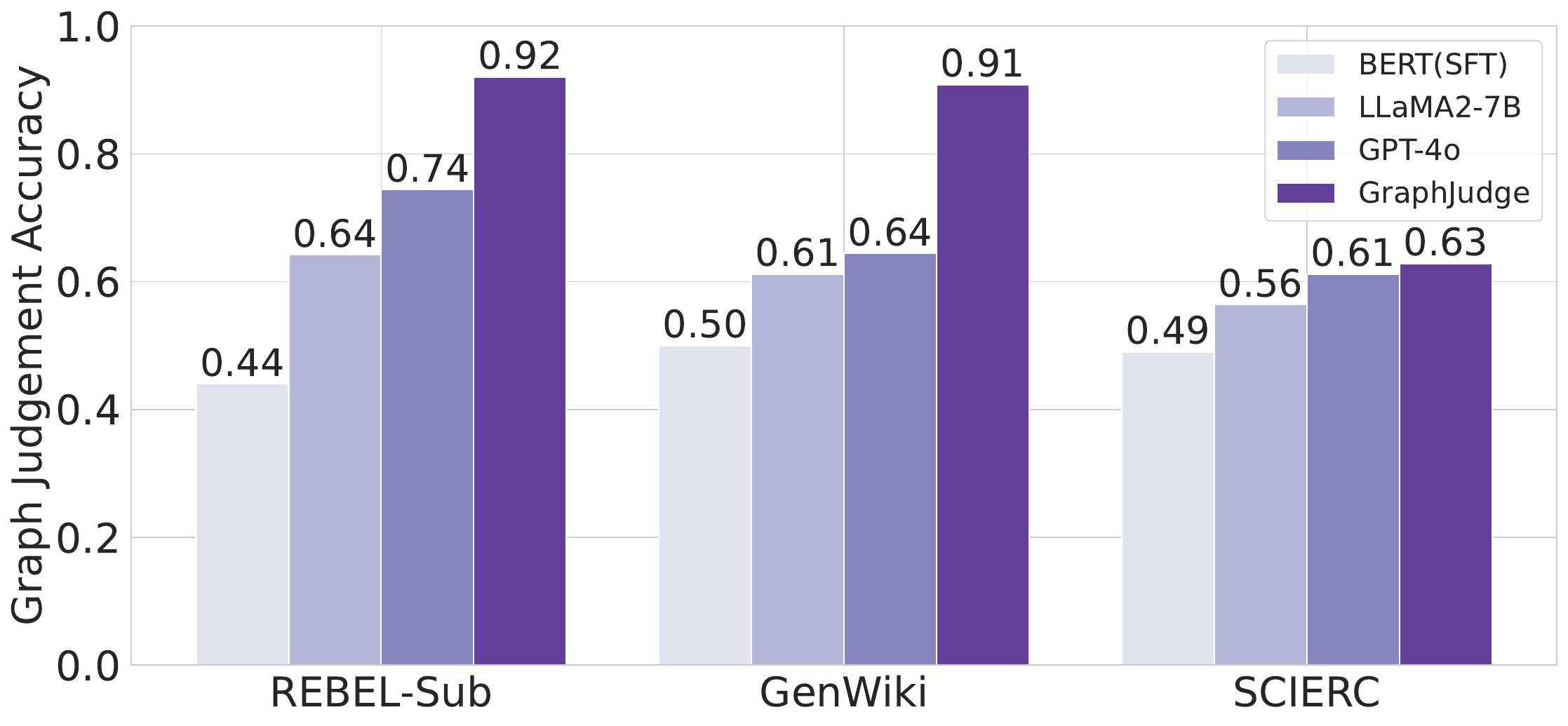}}
\caption{A comparison of the capabilities of bert-base-uncased (SFT)~\cite{devlin2019bert}, LLaMA-2-7B, GPT-4o, and our GraphJudge in graph judgment tasks.}
\label{gj}
\vspace{-0.15in}
\end{figure}
\textbf{Effect of Knowledge Aware Supervised Fine-Tuning}. We conduct graph judgement on the triples without fine-tuning the open-source LLM, which is denoted as `w/o KASFT'. The result in Table~\ref{tab:ablation} indicates that without SFT, the naive LLM has weak graph judgement abilities. And with a fine-tuned LLM as a graph judge, the performance can be improved a lot. Because KASFT enables the LLM to acquire both fact-checking capabilities and domain-specific knowledge within the triples in our instruction training data. 

Furthermore, we apply negative sampling to construct instructions on the \textbf{test set} like what we did on the training set. We randomly select 500 samples and perform graph judgement to compare the capabilities of different models. As shown in Figure~\ref{gj} and Appendix~\ref{appendix:KASFT}, both fine-tuned small models like BERT and naive powerful LLMs like GPT-4o show poor performance on the graph judgement task even with documents as contexts. However, \textbf{GraphJudge can achieve over 90\% judgement accuracy on REBEL-Sub and GenWiki}, which demonstrates the KASFT module can indeed enhance the effectiveness of LLMs as a graph judge.

\begin{table*}[!h]
    \centering
\setlength\tabcolsep{1.6pt}
\renewcommand{\arraystretch}{1.2}
\scalebox{0.85}{
    \begin{tabular}{l|ccc|ccc|ccc}
        \toprule
        \multirow{2}{*}{\textbf{Method}}
        &\multicolumn{3}{c|}{GenWiki @ REBEL-Sub}&\multicolumn{3}{c|}{REBEL-Sub @ GenWiki}&\multicolumn{3}{c}{REBEL-Sub @ SCIERC}\\
        &\textbf{G-BS-F1↑}&\textbf{G-BL-F1↑}&\textbf{G-RO-F1↑}&\textbf{G-BS-F1↑}&\textbf{G-BL-F1↑}&\textbf{G-RO-F1↑}&\textbf{G-BS-F1↑}&\textbf{G-BL-F1↑}&\textbf{G-RO-F1↑}\\
        \midrule
        \textit{GPT-4o}& 0.4163&  0.3158&  0.2966 & 0.8428 & 0.6774 & \textbf{0.6028} & 0.7035 & 0.5032 & 0.4425\\
        \textit{PiVe}&0.4090& 0.3010& 0.2823 & 0.8230 & 0.6503 & 0.5817 & 0.6725 & 0.4924 & 0.4385\\
        \textit{KGGen}&0.4995& 0.3146& 0.2719 & 0.8169 & 0.5542 & 0.4641 & 0.6635 & 0.4725 & 0.4211\\
        \textit{GraphJudge}& \textbf{0.5814}& \textbf{0.4055}& \textbf{0.3649} & \textbf{0.8587} & \textbf{0.6792} & 0.5911 & \textbf{0.7431} & \textbf{0.5156} & \textbf{0.4572}\\
        \bottomrule
    \end{tabular}
}
\caption{Results of generalization study on REBEL-Sub, GenWiki, and SCIERC with the LLM fine-tuned on GenWiki (GenWiki @ REBEL-Sub) and REBEL-Sub (REBEL-Sub @ GenWiki, REBEL-Sub @ SCIERC), respectively.}
\label{tab:generalization}
\setlength{\abovecaptionskip}{3pt}
\vskip -0.1in
\end{table*}
\textbf{Effect of Graph Judgement}. We compare the performances of our full model and the model without GJ module denoted as `w/o GJ'. The result suggests that GJ module plays a very important role in GraphJudge. It can significantly enhance the quality of KGs generated by the closed-source LLM and reduce the effects of the inaccuracies or hallucinations that may arise from LLMs. The closed-source LLM excels in zero-shot generation, boosting recall but suffering accuracy due to hallucinations or knowledge inadequacy. The GJ module relieves this by filtering inaccurate triples, enhancing the quality of constructed KGs.

\subsection{Generalization Capabilities of GraphJudge (RQ3)}
To demonstrate the generalization abilities of GraphJudge, we conduct experiments in cross-dataset scenarios, which are training the LLM on GenWiki and then evaluate it on REBEL-Sub, training the LLM on REBEL-Sub and then evaluate it on GenWiki and SCIERC, respectively. As shown in Table~\ref{tab:generalization}, our method can still outperform baseline methods, which indicates GraphJudge has great capabilities of generalization across various corpus. This is because the ability to check facts with triple structure learned from graph judgement tasks can be generalized. It also suggests that \textbf{GraphJudge once well trained on a general dataset, can be readily applied to diverse datasets with common knowledge}. We also note that there is a gentle performance drop on SCIERC dataset, which is reasonable because there is less in-domain knowledge in the REBEL-Sub dataset than that in SCIERC dataset. And this result can further demonstrate the domain-specific knowledge within the training corpora is usefull in GraphJudge.

\section{Conclusions}
In this paper, we introduce a new method called GraphJudge for automatically constructing KGs, which leverages the potential of LLMs to act as graph judges. In GraphJudge, we propose ECTD, KASFT and GJ modules to mitigate the impact of irrelevant information from documents and exploit the benefits of trainable open-source LLMs and harnessing the strong zero-shot generation capabilities of closed-source LLMs. 
The experiments conducted on two general and one domain-specific datasets demonstrate GraphJudge's consistent superiority against various baseline methods.


\section{Limitations}
GraphJudge has the following limitations. First, even though we employ the LLMs act as both the extractor and the judge to improve the quality of constructed KGs, we still use the entity-level triples to construct KGs and there could be better knowledge units to form a better KG. Second, a more reasonable benchmark to evaluate the quality of constructed KGs should be proposed in the future. Currently, most of the work just utilize the `ground truth' KGs to calculate the correctness and comprehensiveness of constructed KGs, However, the quality of `ground truth' KGs may still deserve suspicion. So using a self-supervised approach to evaluate the KGs is in demands. We will research for more KG constructing and evaluating method to improve the performance of knowledge extraction.

\bibliography{acl_latex}

\appendix

\section*{Appendix}\label{sec:appendix}

\section{Datasets}\label{sec:appendix_dataset}
\begin{table}[h]
	\setlength\tabcolsep{6pt}
	\renewcommand{\arraystretch}{1.2}
	\centering
	\scalebox{0.75}{
		\begin{tabular}{llll}
			\hline
			Dataset    & REBEL-Sub& GenWiki&SCIERC\\
			\hline
			\# of Train KGs& 45,791& 69,788&350\\
			\# of Test KGs& 1,799&1,000 & 100\\
            \hline
 \# of Train Triples& 268,864& 588,642& 6,429\\
			\# of Test Triples& 5,595&3,915&974\\
                \hline
		\end{tabular}
       
	}
    \caption{Statistics of datasets. }
     \label{tab:statistics}
\end{table}

\begin{figure}[thb] \centering
    \includegraphics[width=0.45\textwidth]{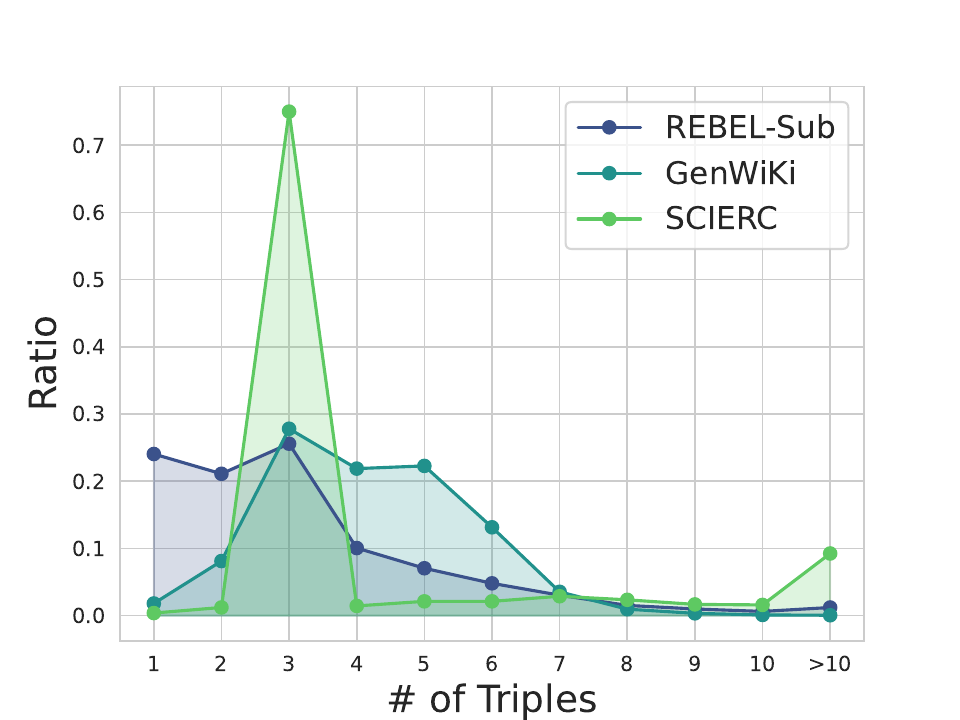}
    \includegraphics[width=0.45\textwidth]{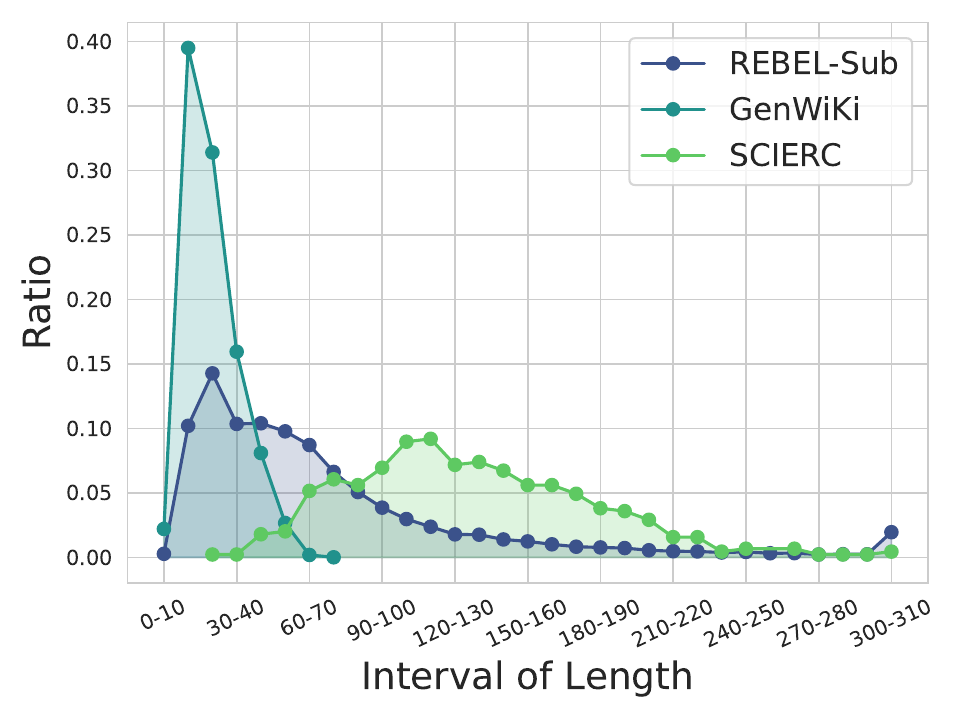}
    \caption{The figure above is the normalized distribution of the number of triplets in each dataset and the figure below is the normalized distribution of the length of documents in each dataset.}
    \label{fig:dataset}
\vspace{-0.15in}
\end{figure}
We conduct experiments on the following three datasets, And we also calculate the percentages of KGs with different numbers of triples and different lengths of original documents in each dataset in Figure~\ref{fig:dataset}. The statistics of each dataset are shown in Table~\ref{tab:statistics}.     

\textbf{REBEL-Sub}. REBEL~\cite{huguet-cabot-navigli-2021-rebel-relation} dataset comes from Wikipedia text before the table of contents, as well as Wikidata for the triplets annotation. The dataset is collected by the extraction pipeline cRocoDiLe~\cite{huguet-cabot-navigli-2021-rebel-relation}. The original REBEL dataset is a large-scale corpus. We utilize a subset of REBEL referred to as REBEL-sub, consisting of 50,000/2,000/2,000 samples for the training, validation, and test set respectively, randomly chosen from the original dataset. Moreover, we filter out the samples with empty ground truth KG.

\textbf{GenWiki}. GenWiki~\cite{jin-etal-2020-genwiki} is an extensive dataset sourced from general Wikipedia, comprising 1.3 million non-parallel texts and graphics that share content. 
In our study, for efficient validation, we use a subset of $\text{GenWiki}_{\texttt{FINE}}$ as our training set and employ another subset of $\text{GenWiki}_{\texttt{FINE}}$ for testing. And the documents in original testing data are too short for us to validate our method. To enhance the quality of training data lacking human annotations, we also exclude the triples with incorrect formats.

\textbf{SCIERC}. SCIERC~\cite{luan2018multitask} is a scientific domain-specific dataset comprises annotations for scientific entities, their relations, and coreference clusters within 500 scientific abstracts. 
It expands upon the datasets from SemEval 2017 Task 10~\cite{augenstein-etal-2017-semeval} and SemEval 2018 Task 7~\cite{gabor-etal-2018-semeval} by introducing additional entity types, relation types, broader relation coverage, and incorporating cross-sentence relations through coreference links. In addition, we filter out the samples with empty ground truth KG.

\textbf{Re-DocRED}. Re-DocRED~\cite{tan2022revisiting} is a well annotated domain-specific dataset that is used in LKD-KGC~\cite{sun2025lkd}. It is a refined version of DocRED~\cite{yao-etal-2019-docred}, enhanced through reannotation of omitted relation triples. We follow the settings in LKD-KGC to use the Windows-centric subset containing domain knowledge about Windows Operation Systems. And because there is no enough samples in this dataset for us to train our model, we use the model trained on GenWiki to evaluate the performance of GraphJudge on this dataset.

\section{Experimental Metrics}\label{sec:appendix_metrics}

In this section, we explain the details of our evaluation metrics.

\textbf{G-BERTScore (G-BS)}: Here we use a matching metric that evaluate the degree of similarity between the ground-truth and predicted graphs, which is called G-BERTScore\cite{saha2021explagraphs}. And it is designed as an extension of the text generation metric BERTScore\cite{zhang2019bertscore}. In G-BERTScore, each triple within knowledge graphs is treated as a sentence, and subsequently, the similarity score between sentences of triples in the ground-truth and predicted knowledge graphs is computed. 
And we compute the accuracy, recall, and F1 score of each constructed KG against the ground-truth using G-BERTScore, denoted as \textbf{G-BS-Acc}, \textbf{G-BS-Recall}, and \textbf{G-BS-F1}, respectively.

\textbf{G-BLEU (G-BL)}: BLEU (Bilingual Evaluation Understudy)\cite{papineni2002bleu} is a metric for evaluating the quality of text which has been machine-translated from one natural language to another. Here we use this approach to determine the resemblance between the triple sentences in the ground-truth and predicted KGs, which is called G-BLEU. The formulas are shown in \eqref{eq:16}, \eqref{eq:17}, \eqref{eq:18}, \eqref{eq:19}, \eqref{eq:20}. $N$ is the maximum order of n-grams considered in the evaluation and we set $N=4$, which is a default number in the Python package\footnote{https://pypi.org/project/bert-score/}. $\text{BP}$ is the brevity penalty, which is used to avoid giving too much credit to short translations. And we compute the accuracy, recall, and F1 score of each constructed KG against the ground-truth using G-BLEU, denoted as \textbf{G-BL-Acc}. \textbf{G-BL-Recall}, and \textbf{G-BL-F1}, respectively.

\begin{equation}
\begin{split}
&\text{Match}(n) =  \\ 
&\quad \sum_{\mathcal{X}_t \in \mathcal{X}_\mathcal{T}} \sum_{\text{gram}(n) \in \mathcal{X}_{t}}  \text{Count}(\text{gram}(n), \mathcal{X}_{\hat{t}}),
\end{split}
\label{eq:18}
\end{equation}
\begin{equation}
\begin{split}
&\text{Total}(n) = \\ 
&\quad \sum_{\mathcal{X}_t \in \mathcal{X}_\mathcal{T}}\sum_{\text{gram}(n)\in \mathcal{X}_{\hat{t}}} \text{Count}(\text{gram}(n), \mathcal{X}_{\hat{t}}),
  \end{split}
    \label{eq:19}
\end{equation}
\begin{equation}
\text{BP} = \begin{cases}
1 & \text{ if }|\mathcal{X}_{\hat{t}}| > |\mathcal{X}_{t}| \\
e^{(1 - \frac{|\mathcal{X}_{t}|}{|\mathcal{X}_{\hat{t}}|})} & \text{ if } |\mathcal{X}_{\hat{t}}| \leq |\mathcal{X}_{t}|,
\end{cases}
    \label{eq:20}
\end{equation}
\begin{equation}
w_n = \frac{\text{Match}(n)}{\text{Total}(n)},
    \label{eq:17}
\end{equation}
\begin{equation}
\text{G-BLEU} = \text{BP} \times (\prod^{N}_{n = 1}w_n)^{\frac{1}{N}}.
    \label{eq:16}
\end{equation}
\textbf{G-ROUGE (G-RO)}: ROUGE (Recall-Oriented Understudy for Gisting Evaluation)\cite{lin-2004-rouge} is a set of metrics for evaluating automatic summarization and machine translation systems. And here we utilize ROUGE to compare the similarities between the triple sentences in the ground-truth and predicted KGs, which is G-ROUGE. Here our G-ROUGE score is based on the notion of n-gram co-occurrence statistics and we set $n=2$. For G-ROUGE-N, which focuses on the overlap of n-grams between the ground-truth triple sentence and the predicted triple sentence, the formulas are shown in \eqref{eq:13}, \eqref{eq:14}, \eqref{eq:15}. Unlike G-BLEU, G-ROUGE is computed using recall as a metric. And $\text{Count}(\text{gram}(n), \mathcal{X}_{\hat{t}})$ is the number of times the n-gram appears in the predicted triple sentence $\mathcal{X}_{\hat{t}}$. And we compute the accuracy, recall, and F1 score of each constructed KG against the ground-truth using G-ROUGE, denoted as \textbf{G-RO-Acc}, \textbf{G-RO-Recall}, and \textbf{G-RO-F1}, respectively.
\begin{equation}
\begin{split}
&\text{Match}(n) = \\
&\quad \sum_{\mathcal{X}_t \in \mathcal{X}_\mathcal{T}}\sum_{\text{gram}(n)\in \mathcal{X}_{t}}\text{Count}(\text{gram}(n), \mathcal{X}_{\hat{t}}),
\end{split}
    \label{eq:13}
\end{equation}
\begin{equation}
\begin{split}
&\text{Total}(n) =  \\
&\quad \sum_{\mathcal{X}_t \in \mathcal{X}_\mathcal{T}}\sum_{\text{gram}(n)\in \mathcal{X}_{t}}\text{Count}(\text{gram}(n), \mathcal{X}_{t}),
\end{split}
    \label{eq:14}
\end{equation}
\begin{equation}
\text{G-ROUGE} = \frac{\text{Match}(n)}{\text{Total}(n)}.
    \label{eq:15}
\end{equation}
\section{Experimental Settings}\label{sec:appendix_settings}
\begin{table}[h]
	\setlength\tabcolsep{2pt}
	\renewcommand{\arraystretch}{1}
	\centering
	\scalebox{0.9}{
		\begin{tabular}{cc}
			\hline
			Hyper-parameter    & Experimental Setting \\
			\hline
			Micro Batch Size& 8\\
			Batch Size& 128\\
			Gradient Accumulation Steps& 16\\
			Training Steps& 500\\
			Learning Rate& 3e-4\\
			Lora Attention Dimension& 8\\
			Alpha Parameter& 16\\
			Target Modules& q\_proj, v\_proj\\
 Warmup Steps&100\\
			Optimizer& AdamW\\
			\hline 
		\end{tabular}
		
	}
    \caption{Implementation detail of SFT in GraphJudge. }
    \label{tab:hyper_parameter}
\end{table}
During the Knowledge Aware Supervised Fine-Tuning module, we follow the parameter settings in Table~\ref{tab:hyper_parameter} referring to the tuning process used for triple classification tasks in the KG-LLaMA~\cite{yao2023exploring}.

\section{Graph Judgement Algorithm}\label{sec:graphjudgement_algorithm}
\begin{algorithm}[!h]
	\renewcommand{\algorithmicrequire}{\textbf{Input:}}
	\renewcommand{\algorithmicensure}{\textbf{Output:}}
	\caption{The Graph Judgement procedure of GraphJudge}
    \label{algorithm:gj}
	\begin{algorithmic}[1]  
		\REQUIRE  The fine-tuned expert LLM $p_{\theta}$ ; Candidate triples $\mathcal{T}^{*}$ in the draft KGs $\mathcal{S}_{\mathcal{G}^{*}}$; Paired refined text $\mathcal{D}^{*}$ of the candidate triples;
		\ENSURE  The predicted KGs $\mathcal{S}_{\hat{\mathcal{G}}}$ with refined triples $\hat{\mathcal{T}}$;
        \STATE $\mathcal{S}_{\hat{\mathcal{G}}} \leftarrow \{\}$;
        \STATE $\hat{\mathcal{T}} \leftarrow \{\}$;
		\FOR{\textit{each $\mathcal{G}^{*}$ in $\mathcal{S}_{\mathcal{G}^{*}}$,\\ each denoised document $ d^{*} $ in $\mathcal{D}^{*}$}}  
		\STATE $\hat{\mathcal{R}} \leftarrow \{\}$;
        \STATE $\hat{\mathcal{E}} \leftarrow \{\}$;
        \FOR{\textit{each triple $t^{*} = <h, r, t> \in \mathcal{G}^{*}$}}
        \STATE \textit{/*Transform the triple and refined text into a sentence*/}
        \STATE $\mathcal{X}_{t^{*}} = \text{Sentence}(<h,r,t>, d^{*})$;
        \STATE \textit{/*Verify the correctness of the current triple with fine-tuned LLM*/}
        \STATE $\hat{y}_{t^{*}} = \text{Pred}(p_{\theta}(\mathcal{X}_{t^{*}}))$;
        \IF{\textit{$\hat{y}_{t^{*}}$ is not `False'}}
        \STATE $\hat{\mathcal{T}} \leftarrow t^{*}$;
        \STATE $\hat{\mathcal{R}} \leftarrow \hat{\mathcal{R}} \cup \{r\}$;
        \STATE $\hat{\mathcal{E}} \leftarrow \hat{\mathcal{E}} \cup \{h,t\}$;
        \ENDIF
        \ENDFOR
        \STATE $\hat{\mathcal{G}} = \{<h,r,t> | h, t \in \hat{\mathcal{E}}, r \in \hat{\mathcal{R}}, <h,r,t> \in \hat{\mathcal{T}}\}$;
        \STATE $\mathcal{S}_{\hat{\mathcal{G}}} \leftarrow \mathcal{S}_{\hat{\mathcal{G}}} \cup \{\hat{\mathcal{G}}\}$;
\ENDFOR
	\end{algorithmic}
\end{algorithm}
The detailed procedure of the graph judgement algorithm is demonstrated in Algorithm~\ref{algorithm:gj}.

\begin{figure*}[h]
\centerline{\includegraphics[width=0.8\linewidth]{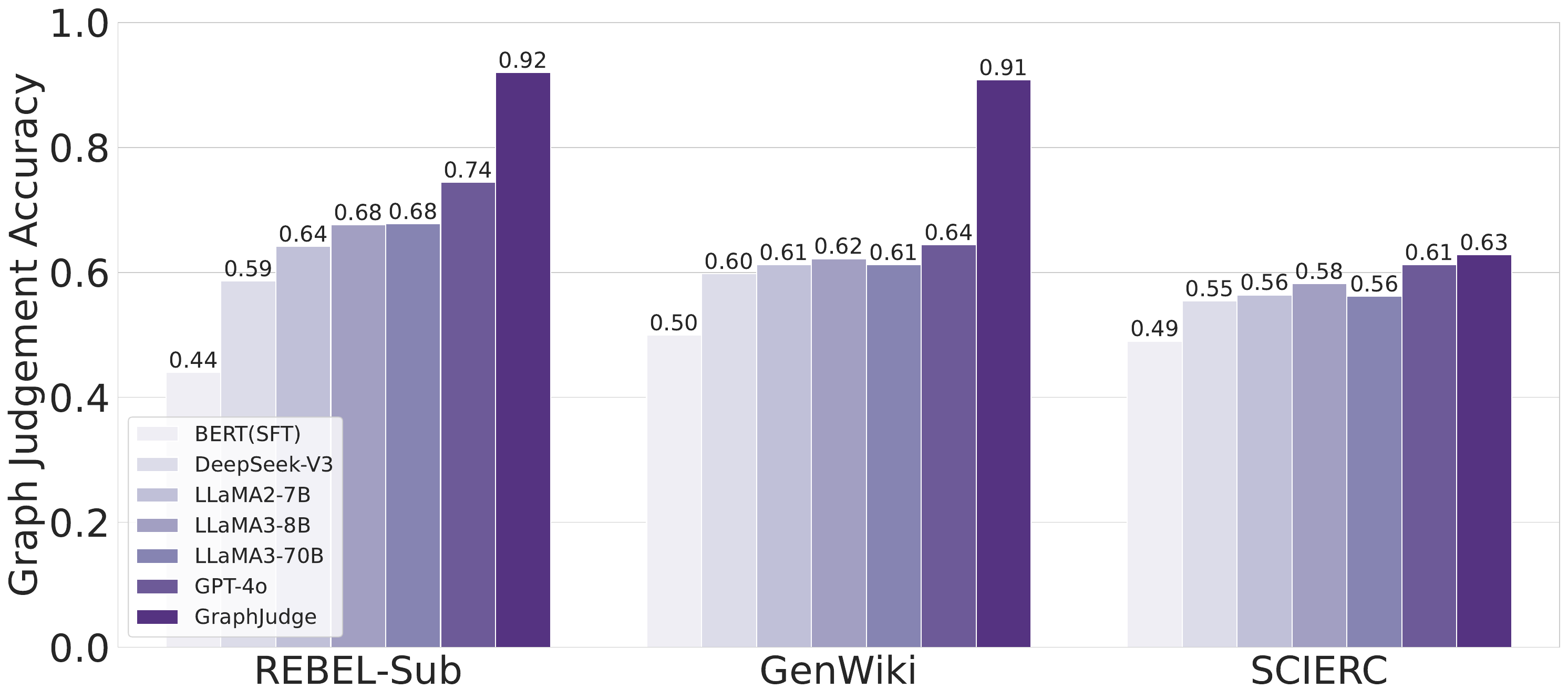}}
\caption{A comparison of the capabilities of bert-base-uncased (SFT)~\cite{devlin2019bert}, DeepSeek-V3~\cite{liu2024deepseek}, LLaMA-2-7B, LLaMA-3-8B, LLaMA-3-70B, GPT-4o, and our GraphJudge in graph judgment tasks.}
\label{more_gj}
\vskip -0.1in
\end{figure*}
\begin{table}[h]
\centering
\small
\vspace*{-0.25cm}
\setlength\tabcolsep{4pt}
\renewcommand{\arraystretch}{1.2}
 \resizebox{0.46\textwidth}{!}{\begin{tabular}{c|c|cc}
\toprule
\multirow{2}{*}{\textbf{Dataset}} &
\multirow{2}{*}{\textbf{Context}} &
\multicolumn{2}{c}{\textbf{Model}} \\
\cmidrule(lr){3-4}
     & & \textbf{LLaMA-3-8B}& \textbf{LLaMA-3-70B}\\ \midrule
    
\multirow{3}{*}{REBEL-Sub}& [Lower-Upper]& 47.33-99.33& 76.67-100.0\\
                            & $\mathcal{D}^{*}$& 94.67& 96.00\\
                            & $\hat{\mathcal{G}}$& 85.33& 90.67\\
    \midrule
\multirow{3}{*}{GenWiki}& [Lower-Upper]& 53.33-100.0& 68.00-100.0\\
                            & $\mathcal{D}^{*}$& 98.00& 96.00\\
                            & $\hat{\mathcal{G}}$& 93.00& 93.00\\
    \midrule
\multirow{3}{*}{SCIERC}& [Lower-Upper]& 65.00-99.67& 77.00-99.67\\
                            & $\mathcal{D}^{*}$& 95.67& 96.33\\
                            & $\hat{\mathcal{G}}$& 90.33& 93.67\\
\bottomrule
\end{tabular}
}
\caption{MCQ performance across datasets. Each row displays the lower-upper bound performance (no context vs. original document), denoised document performance, and our KG performance for different models. Using $\mathcal{D}^{*}$ and $\hat{\mathcal{G}}$ preserves most information for answering MCQs, perform close to the using the original document (upper bound) across datasets and models.}
\label{tab:mcq_result}
\vskip -0.1in
\end{table}

\section{Entity Coverage of the LLM Extraction}\label{appendix:entity_coverage}
\begin{table}[h]
    \centering
    \small
    \begin{tabular}{l|l|cc}
    \toprule
    Dataset & Method & Recall & Acc \\
    \midrule
    \multirow{2}{*}{REBEL-Sub} & \textit{w/o GJ} & 0.9784 & 0.3709 \\
                               & \textit{w/ GJ}  & 0.9411 & 0.5495 \\
    \midrule
    \multirow{2}{*}{GenWiki-Hard} & \textit{w/o GJ} & 0.9365 & 0.7932 \\
                                  & \textit{w/ GJ}  & 0.9018 & 0.8343 \\
    \midrule
    \multirow{2}{*}{SCIERC} & \textit{w/o GJ} & 0.9392 & 0.6916 \\
                            & \textit{w/ GJ}  & 0.9187 & 0.7176 \\
    \bottomrule
    \end{tabular}
    \caption{Entity coverage and accuracy across datasets.}
    \label{tab:entity_coverage}
\vspace{-0.15in}
\end{table}
To verify the comprehensiveness of extracted entities by the LLM, we evaluate the entity coverage as well as accuracy across three datasets. In detail, we use the semantic metric of Recall and Accuracy based on BertScore to measure the entity coverage and accuracy, similar to the metric G-BS described in the paper. As shown in Table~\ref{tab:entity_coverage}, we report the coverage as well as accuracy of the extracted entities before (w/o GJ) and after (w/ GJ) the GJ module. And here are the insights: (1) The entities extracted from the original text have high coverage and low accuracy, which verifies that \textbf{the extracted entities can provide good foundation for both denoising and relation extraction}. (2) The entities within the triples filtered by GJ module still keep a high coverage and have a significantly higher accuracy. It further validates the effectiveness of our GJ module.

\section{Effect of ECTD Module}~\label{appendix:ECTD}
To validate that ECTD module can lead to a cleaner refined text, we sample a text-graph pair from the REBEL-Sub dataset. Then we split the original and refined document into the same number of chunks, which we set $20$ here. And we use a PLM BERT (bert-base-uncased\footnote{https://huggingface.co/google-bert/bert-base-uncased})~\cite{devlin2019bert} to process these chunks and get the embedding of each chunk. And we calculate the cosine similarities between these document chunks and triple sentences, as shown in Figure~\ref{relavance}. Deeper color in the heat maps suggests a stronger relevance between the specific triple and document chunk.

\section{Knowledge Retention of ECTD Module and KG}\label{appendix:ECTD_know}
While the ECTD module has the ability to remove irelevant information contained in the original documents, it is also necessary to verify that the important knowledge is well preserved in the documents denoised by ECTD. We test how well \textbf{multiple-choice question}\textbf{ (MCQ)} performance is preserved after we refined the original documents. 

In detail, similar to the existing work~\cite{schuhmann2025project}, we generate various MCQs with LLaMA-3-70B for each original document. For REBEL-Sub, we randomly sample 500 documents and generate 3 MCQs for each document. For SCIERC, because the test set of that is very small, we used the full test set of SCIERC with 3 MCQs for each document. For GenWiki, because the average lengths of the documents are very short, we generate only 1 MCQ for each document. Then we ask LLaMA-3-8B to answer them with no context (denoted as lower bound), then ask them again with the original passage (denoted as upper bound) for sanity check. Finally, we conduct tests using denoised documents (denoted as $\mathcal{D}^{*}$) and KG triples constructed by GraphJudge (denoted as $\hat{\mathcal{G}}$). The results in Table~\ref{tab:mcq_result} demonstrates that MCQs performance with $\mathcal{D}^{*}$ or $\hat{\mathcal{G}}$ remains far above the lower bound baseline and approaches the original-document upper bound. It proofs that \textbf{important information is well preserved in both our denoised documents and constructed KG.}

\section{Effect of KASFT Module}\label{appendix:KASFT}
In this section, we extend the baseline models to explore their abilities to be a graph judge with the same experimental settings in Section~\ref{sec:ablation}. As shown in Figure~\ref{more_gj}, We extend baseline models to fine-tuned BERT, DeepSeek-V3, LLaMA-2-7B, LLaMA-3-8B, LLaMA-3-70B, and GPT-4o. Compared with them, our proposed GraphJudge demonstrates consistent superiority in graph judgement tasks, which further proofs that the KASFT module can improve the capabilities of open-source LLMs as a graph judge. And neither fine-tuning a PLM with a smaller parameter size nor directly employing a powerful closed-source LLM can achieve a high accuracy on graph judgement tasks, which suggests the necessity to introduce our proposed GraphJudge.

\begin{figure*}[h] 
    \begin{AIbox}{Prompt Template for Graph Judgement.}
    Goal:\\
    You need to do the graph judgement task, which means you need to clarify the correctness of the given triples with the given original document.\\
    Here is the question:\\
    According to the original document: {\color{deepblue}\bf \{text\}} \\
    Is this true: {\color{deepblue}\bf \{head\_entity\}} {\color{deepblue}\bf \{relation\}} {\color{deepblue}\bf \{tail\_entity\}}?
    \tcblower
    {\bf \large Output:}\\
    No, it is not true./ Yes, it is true.
    \end{AIbox}
    \vspace{-1em}
    \caption{The prompt template for the open-source LLM LLaMA to construct graph judgement instructions.}
    \label{gj_prompt}
\end{figure*}

\begin{figure*}[!ht] 
    \begin{AIbox}{Prompt Template for MCQ Answering.}
    Given the context or evidence:\\
    {\color{deepblue}\bf \{context\}} \\
    Here is a multiple-choice question:\\
    Question:\\
    {\color{deepblue}\bf \{question\}} \\
    Options:\\
    A. {\color{deepblue}\bf \{option\_A\}} \\
    B. {\color{deepblue}\bf \{option\_B\}} \\
    C. {\color{deepblue}\bf \{option\_C\}} \\
    D. {\color{deepblue}\bf \{option\_D\}} \\
    Please select the correct answer by choosing A, B, C, or D. Respond with only the letter of your choice.
    \tcblower
    {\bf \large Output:}\\
    A/B/C/D
    \end{AIbox}
    \vspace{-1em}
    \caption{The prompt template for the MCQ answering.}
    \label{mcq_ans_prompt}
\end{figure*}

\begin{figure*}[!ht] 
    \begin{AIbox}{Prompt Template for Entity Extraction.}
    Goal:\\
    Transform the text into a list of entities. Please ensure the comprehensiveness and accuracy of the extracted entities, which should be related to the topic of the text.\\
    
    Here are two examples:\\
    Example\#1:\\
    Text: "Shotgate Thickets is a nature reserve in the United Kingdom operated by the Essex Wildlife Trust."\\
    List of entities: ["Shotgate Thickets", "Nature reserve", "United Kingdom", "Essex Wildlife Trust"]\\
    Example\#2:\\
    Text: "Garczynski Nunatak is a cone-shaped nunatak, the highest in a cluster of nunataks close west of Mount Brecher, lying at the north flank of Quonset Glacier in the Wisconsin Range of the Horlick Mountains of Antarctica."\\
    List of entities: ["Garczynski Nunatak", "nunatak", "Wisconsin Range", "Mount Brecher", "Quonset Glacier", "Horlick Mountains", "Antarctica"]\\
    
    Refer to the examples and here is the question:\\
    Text: {\color{deepblue}\bf \{text\}} \\
    List of entities: \\
    \tcblower
    {\bf \large Output:}
    \begin{lstlisting}[style=prompt]
    [{entity_1}, {entity_2}, ... , {entity_n}]
    \end{lstlisting}
    \end{AIbox}
    \vspace{-1em}
    \caption{The prompt template for the closed-source LLM GPT-4o-mini to extract entities.}
    \label{entity_prompt}
\end{figure*}

\begin{figure*}[!ht] 
    \begin{AIbox}{Prompt Template for Relation Extraction.}
    Goal:\\
    Transform the text into a semantic graph(a list of triples) with the given text and entities. Extract subject-predicate-object triples from the assistant message. A predicate (1-3 words) defines the relationship between the subject and object. Relationship may be fact or sentiment based on assistant’s message. Subject and object are entities. Entities provided are from the assistant message and prior conversation history, though you may not need all of them. This is for an extraction task, please be thorough, accurate, and faithful to the 
    reference text.\\

    Note:\\
    1.Generate triples as many as possible. \\
    2.Make sure each item in the list is a triple with strictly three items.\\
    Here are two examples:\\
    Example\#1: \\
    Text: ``Shotgate Thickets is a nature reserve in the United Kingdom operated by the Essex Wildlife Trust.''\\
    Entity List: [``Shotgate Thickets'', ... , ``Essex Wildlife Trust'']\\
    Semantic Graph: [[\"Shotgate Thickets\", ``instance of'', ``Nature reserve''], ...]\\
    Example\#2:\\
    Text: ..\\
    Semantic Graph: ..\\
    Refer to the examples and here is the question:\\
    Text: {\color{deepblue}\bf \{text\}} \\
    Entity List: {\color{deepblue}\bf \{entities\}} \\
    Semantic graph: \\
    \tcblower
    {\bf \large Output:}\\
    \begin{lstlisting}[style=prompt]
    ```
    [
        {triple_1},
        {triple_2},
        ...
        {triple_n}
    ]
    ```
    \end{lstlisting}
    \end{AIbox}
    \vspace{-1em}
    \caption{The prompt template for the closed-source LLM GPT-4o-mini to construct KGs.}
    \label{re_prompt}
\end{figure*}

\begin{figure*}[!ht] 
    \begin{AIbox}{Prompt Template for MCQ Generation.}
    You are an expert in generating multiple-choice questions (MCQs) from scientific text. Your task is to generate {\color{deepblue}\bf \{n\}} MCQs based on the following document:\\
    Each question should:\\
    - Focus on the factual claims, numerical data, definitions, or relational knowledge from the document.\\
    - Have 4 options (one correct and three plausible distractors).\\
    - Clearly indicate the correct answer.\\
    \\
    The output should be in JSON format, with each question as a dictionary containing:\\
    - ``question'': The MCQ question.\\
    - ``options'': A list of 4 options (e.g., [``A: ...'', ``B: ...'', ``C: ...'', ``D: ...'']).\\
    - ``answer'': The correct answer (e.g., ``A'').\\

    Output Example:
    \begin{lstlisting}[style=prompt]
    ```
    [
        {
            "question": "What is the primary role of a catalyst in a chemical reaction?",
            "options": [
                "A": "A catalyst is a substance that increases the rate of a chemical reaction without being consumed in the reaction.",
                "B": "A catalyst is a substance that decreases the rate of a chemical reaction without being consumed in the reaction.",
                "C": "A catalyst is a substance that is consumed in the reaction.",
                "D": "A catalyst is a substance that is not consumed in the reaction."
            ],
            "answer": "A"
        },
        ...
    ]
    ```
    \end{lstlisting}
    Passage:\\
    {\color{deepblue}\bf \{passage\}}
    \tcblower
    {\bf \large Output:}
    \begin{lstlisting}[style=prompt]
    ```
    [
        {MCQ_1},
        {MCQ_2},
        ...
        {MCQ_n}
    ]
    ```
        \end{lstlisting}
    \end{AIbox}
    \vspace{-1em}
    \caption{The prompt template for the MCQ generations.}
    \label{mcq_gen_prompt}
\end{figure*}

\section{Prompt Templates}\label{appendix:prompt_template}
As shown in Figure~\ref{gj_prompt} and Figure~\ref{re_prompt}, we demonstrate the prompt templates for the closed-source LLM to conduct relation extraction and for the open-source LLM to perform graph judgements on the results generated from the closed-source LLM. We also provide the prompt templates used to generate and answer MCQs in Appendix~\ref{appendix:ECTD_know}.
\section{Case Study}
In this section, we present an instance of constructing a KG from a document, achieved through the integration of a naive LLM (GPT-4o-mini) and our GraphJudge. We select a text-graph pair from the SCIERC dataset and contrast the results yielded by our approach with that of GPT-4o-mini. As shown in Table~\ref{tab:case_1}, the KG constructed by GPT-4o-mini with the given original document includes lots of meaningless triples. For example, \textless \textit{We}, \textit{suggest}, \textit{goal}\textgreater , \textless \textit{We}, \textit{suggest}, \textit{evaluation criterion}\textgreater, \textless \textit{We}, \textit{present}, \textit{measure}\textgreater, \textless \textit{We}, \textit{present}, \textit{selection function}\textgreater, etc. It is obvious that these triples do not convey any beneficial information that could be applied to subsequent tasks. And the triple \textless \textit{evaluation criterion}, \textit{new}, \textit{goal}\textgreater does not even follow the general structure of triples, which means that the adjective word `new' is generally not employed as a relational term within triples. The naive LLM have strong zero-shot ability to generate them but it does not have the capability to determine whether they are useful. However, there are no such triples in the KG constructed by our GraphJudge. On the one hand, this is because the triples without any useful information will be clarified as wrong triples by our fine-tuned LLM in graph judgement module. On the other hand, as demonstrated in the case, the document refined by ECTD module exhibits enhanced standardization and a reduction in irrelevant terms, for instance, terms such as `-LRB-' and `-RRB-' have been excluded as they are irrelevant to the document's subject matter.
\begin{table*}[ht]
    \centering
    \vspace{-0.1in}
    \small
    \begin{tabularx}{1\textwidth}{X X}
        \toprule
         
          \textbf{Original Document:} We suggest a new goal and evaluation criterion for word similarity measures .The new criterion -- meaning entailing substitutability -- fits the needs of semantic-oriented NLP applications and can be evaluated directly -LRB-independent of an application -RRB- at a good level of human agreement. Motivated by this semantic criterion we analyze the empirical quality of distributional word feature vectors and its impact on word similarity results, proposing an objective measure for evaluating feature vector quality. Finally, a novel feature weighting and selection function is presented , which yields superior feature vectors and better word similarity performance.\\
          \midrule
          \textbf{Ground-Truth Knowledge Graph:}\\
         \setlength\intextsep{0pt}
          \begin{wrapfigure}{l}{0.35\linewidth}
            \includegraphics[width=\linewidth]{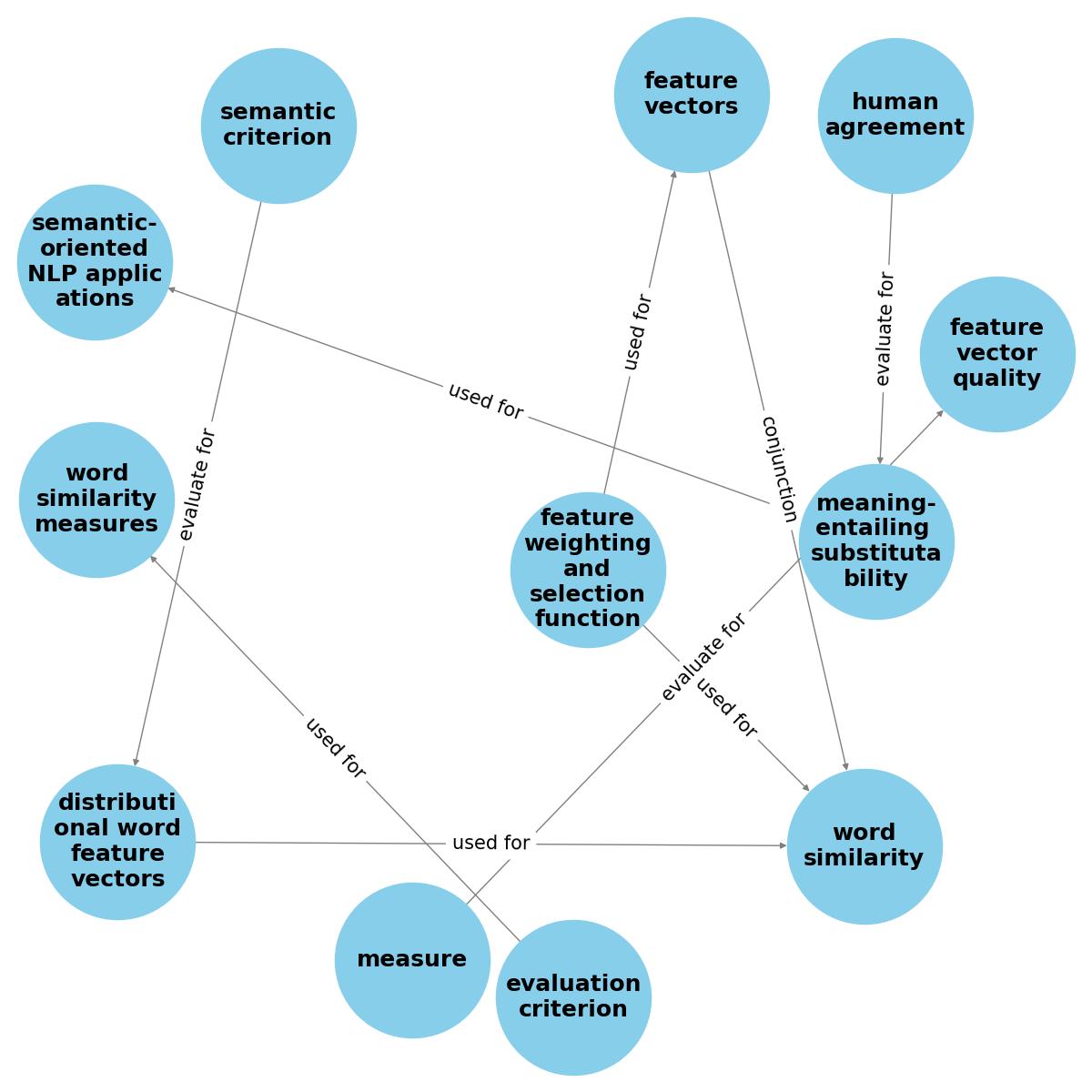}
        \end{wrapfigure}
    [["feature weighting and selection function", "used for", "word similarity"], ["measure", "evaluate for", "feature vector quality"], ["feature vectors", "conjunction", "word similarity"], ["evaluation criterion", "used for", "word similarity measures"], ["meaning-entailing substitutability", "used for", "semantic-oriented NLP applications"],  ["human agreement", "evaluate for", "meaning-entailing substitutability"], ["semantic criterion", "evaluate for", "distributional word feature vectors"], ["distributional word feature vectors", "used for", "word similarity"], ["feature weighting and selection function", "used for", "feature vectors"]]
    
        \\
        \\
        \\
        \\
        \\
        \\
        \\
         \midrule
        \textbf{GPT-4o-mini:} \\
         \setlength\intextsep{0pt}
          \begin{wrapfigure}{l}{0.35\linewidth}
            \includegraphics[width=\linewidth]{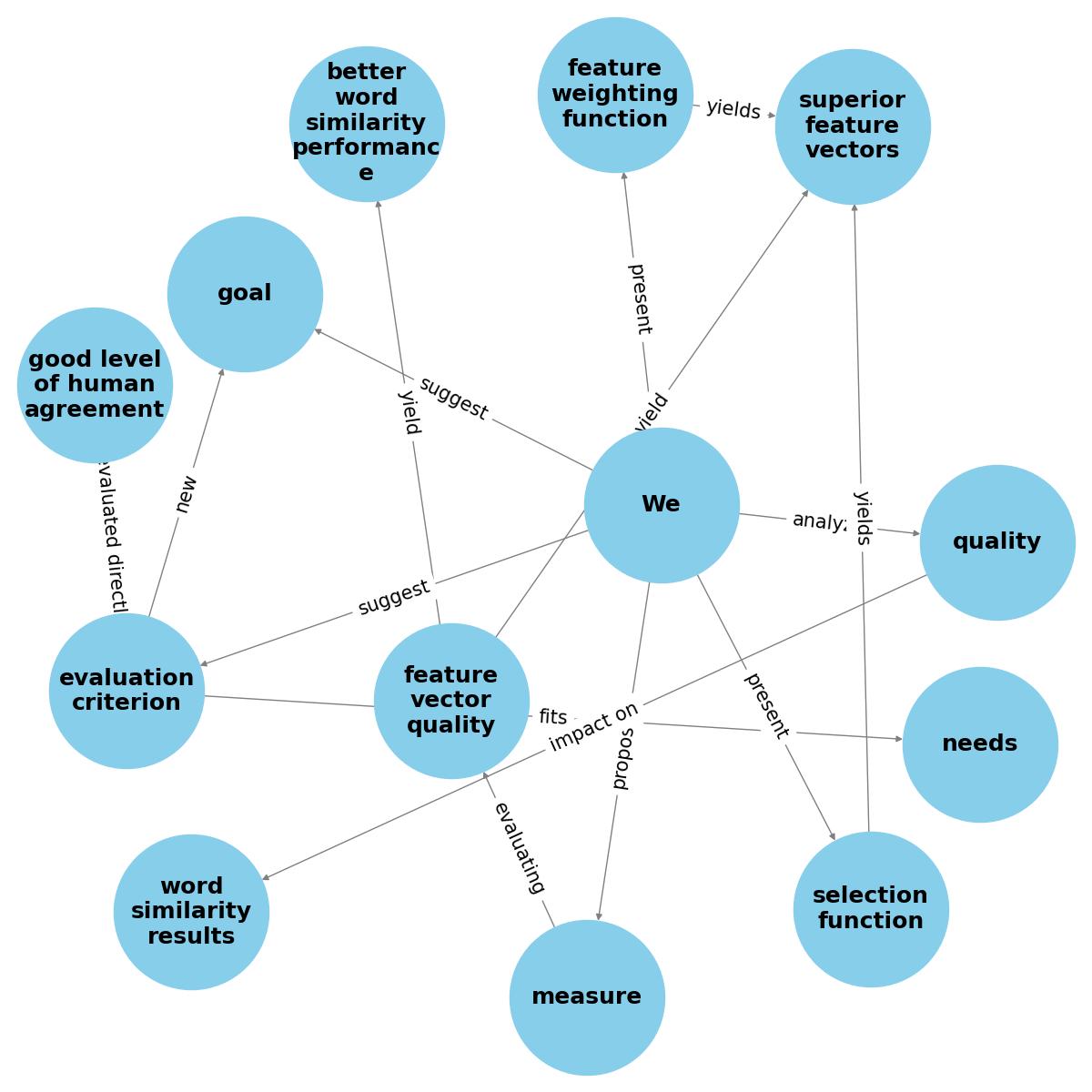}
        \end{wrapfigure}
    [["We", "suggest", "goal"], ["We", "suggest", "evaluation criterion"], ["We", "propose", "measure"], ["evaluation criterion", "new", "goal"], ["evaluation criterion", "fits", "needs"], ["evaluation criterion", "evaluated directly", "good level of human agreement"], ["We", "analyze", "quality"], ["quality", "impact on", "word similarity results"], ["measure", "evaluating", "feature vector quality"], ["feature vector quality", "yield", "superior feature vectors"], ["feature vector quality", "yield", "better word similarity performance"], ["We", "present", "feature weighting function"], ["feature weighting function", "yields", "superior feature vectors"], ["We", "present", "selection function"], ["selection function", "yields", "superior feature vectors"]]
        \\
        \\
        \\
        \\
        \\
        \\
        \\
          \midrule
          
          \textbf{GraphJudge:} \\
         \setlength\intextsep{0pt}
          \begin{wrapfigure}{l}{0.35\linewidth}
            \includegraphics[width=\linewidth]{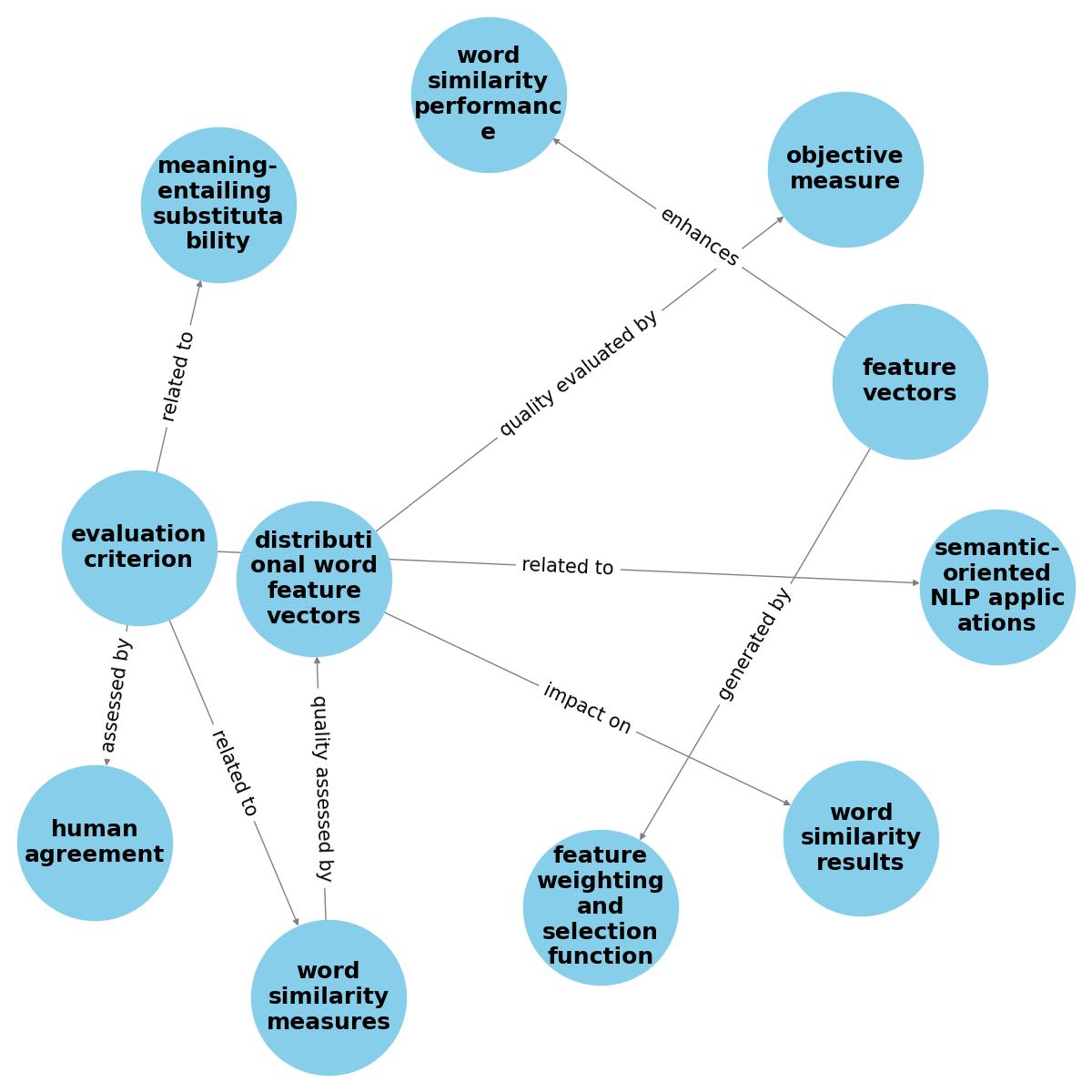}
        \end{wrapfigure}
         [['evaluation criterion', 'related to', 'word similarity measures'], ['evaluation criterion', 'assessed by', 'human agreement'], ['evaluation criterion', 'related to', 'semantic-oriented NLP applications'], ['evaluation criterion', 'related to', 'meaning-entailing substitutability'], ['word similarity measures', 'quality assessed by', 'distributional word feature vectors'], ['distributional word feature vectors', 'impact on', 'word similarity results'], ['distributional word feature vectors', 'quality evaluated by', 'objective measure'], ['feature vectors', 'generated by', 'feature weighting and selection function'], ['feature vectors', 'enhances', 'word similarity performance']]
    
    ~
    
        \textbf{Denoised Document}: We propose a new evaluation criterion for word similarity measures, called meaning-entailing substitutability, which is suitable for semantic-oriented NLP applications. This criterion can be assessed independently of any specific application with a high level of human agreement. We examine the quality of distributional word feature vectors and their impact on word similarity results, introducing an objective measure for evaluating the quality of these feature vectors. Additionally, we present a novel feature weighting and selection function that generates superior feature vectors and enhances word similarity performance.
        \\
         \bottomrule
    \end{tabularx}
    \caption{Comparison of Construction Results between our GraphJudge and GPT-4o-mini.}
      \label{tab:case_1}
    \vspace{-0.15in}
\end{table*}
\end{document}